\documentclass[10pt,journal,compsoc]{IEEEtran}

\usepackage{amsmath,amsfonts}
\usepackage{algorithmic}
\usepackage{algorithm}
\usepackage{array}
\usepackage[caption=false,font=normalsize,labelfont=sf,textfont=sf]{subfig}
\usepackage{textcomp}
\usepackage{stfloats}
\usepackage{url}
\usepackage{verbatim}
\usepackage{graphicx}
\usepackage{cite}
\usepackage{hyperref}
\usepackage{ragged2e}

\usepackage{multirow}
\usepackage{booktabs}
\usepackage{newtxtext,newtxmath}
\usepackage[table]{xcolor}
\usepackage{xcolor}
\usepackage{marvosym}

\begin{document}

\title{UniVST: A Unified Framework for Training-free Localized Video Style Transfer}

\author{
Quanjian Song,
Mingbao Lin,
Wengyi Zhan,
Shuicheng Yan,~\IEEEmembership{Fellow,~IEEE}, \\
Liujuan Cao\,\Letter,
Rongrong Ji,~\IEEEmembership{Senior Member,~IEEE}

\IEEEcompsocitemizethanks{
    \IEEEcompsocthanksitem Q. Song, W. Zhan and L. Cao (Corresponding  Author) are with the Key Laboratory of Multimedia Trusted Perception and Efficient Computing, Ministry of Education of China, Xiamen University, China. (email: \{songqj, zhanwy\}@stu.xmu.edu.cn, caoliujuan@xmu.edu.cn)\protect 
    
    \IEEEcompsocthanksitem M. Lin is with the Rakuten Asia Pte. Ltd., Singapore 048946. (e-mail: linmb001@outlook.com)\protect
    \IEEEcompsocthanksitem S. Yan is with the School of Computing, National University of Singapore. (e-mail: shuicheng.yan@gmail.com)\protect
    
    \IEEEcompsocthanksitem R. Ji is with the Key Laboratory of Multimedia Trusted Perception and Efficient Computing, Ministry of Education of China, Xiamen University, China, also with Institute of Artificial Intelligence, Xiamen University, Xiamen 361005, China. (e-mail: rrji@xmu.edu.cn)\protect
}

}

\IEEEtitleabstractindextext{
\begin{abstract}
\justifying
This paper presents UniVST, a unified framework for localized video style transfer based on diffusion models. It operates without the need for training, offering a distinct advantage over existing diffusion methods that transfer style across entire videos. The endeavors of this paper comprise:
(1) A point-matching mask propagation strategy that leverages the feature maps from the DDIM inversion. This streamlines the model's architecture by obviating the need for tracking models. 
(2) A training-free AdaIN-guided localized video stylization mechanism that operates at both the latent and attention levels. This balances content fidelity and style richness, mitigating the loss of localized details commonly associated with direct video stylization. 
(3) A sliding-window consistent smoothing scheme that harnesses optical flow within the pixel representation and refines predicted noise to update the latent space. This significantly enhances temporal consistency and diminishes artifacts in stylized video.
Our proposed UniVST has been validated to be superior to existing methods in quantitative and qualitative metrics. It adeptly addresses the challenges of preserving the primary object's style while ensuring temporal consistency and detail preservation.
Our code is available at \url{https://github.com/QuanjianSong/UniVST}.
\end{abstract}

\begin{IEEEkeywords}
Localized Video Style Transfer, Diffusion Models, Training-free.
\end{IEEEkeywords}
}

\maketitle

\section{Introduction}
Video editing has greatly improved thanks to the use of diffusion models~\cite{Flatten, Ground-A-Video, rerender-a-video}. 
T2V-Zero~\cite{T2V-Zero} changes self-attention mechanisms to cross-frame attention. Tune-A-Video customizes video editing by adjusting attention weights~\cite{Tune-A-Video}. Fate-Zero~\cite{Fate-Zero} keeps the video content intact via information from the inversion process. Animate-Zero~\cite{Animate-Zero} and VideoBooth~\cite{VideoBooth} add layers that stress time.

\begin{figure}[t]
\centering
\includegraphics[width=0.95\linewidth]{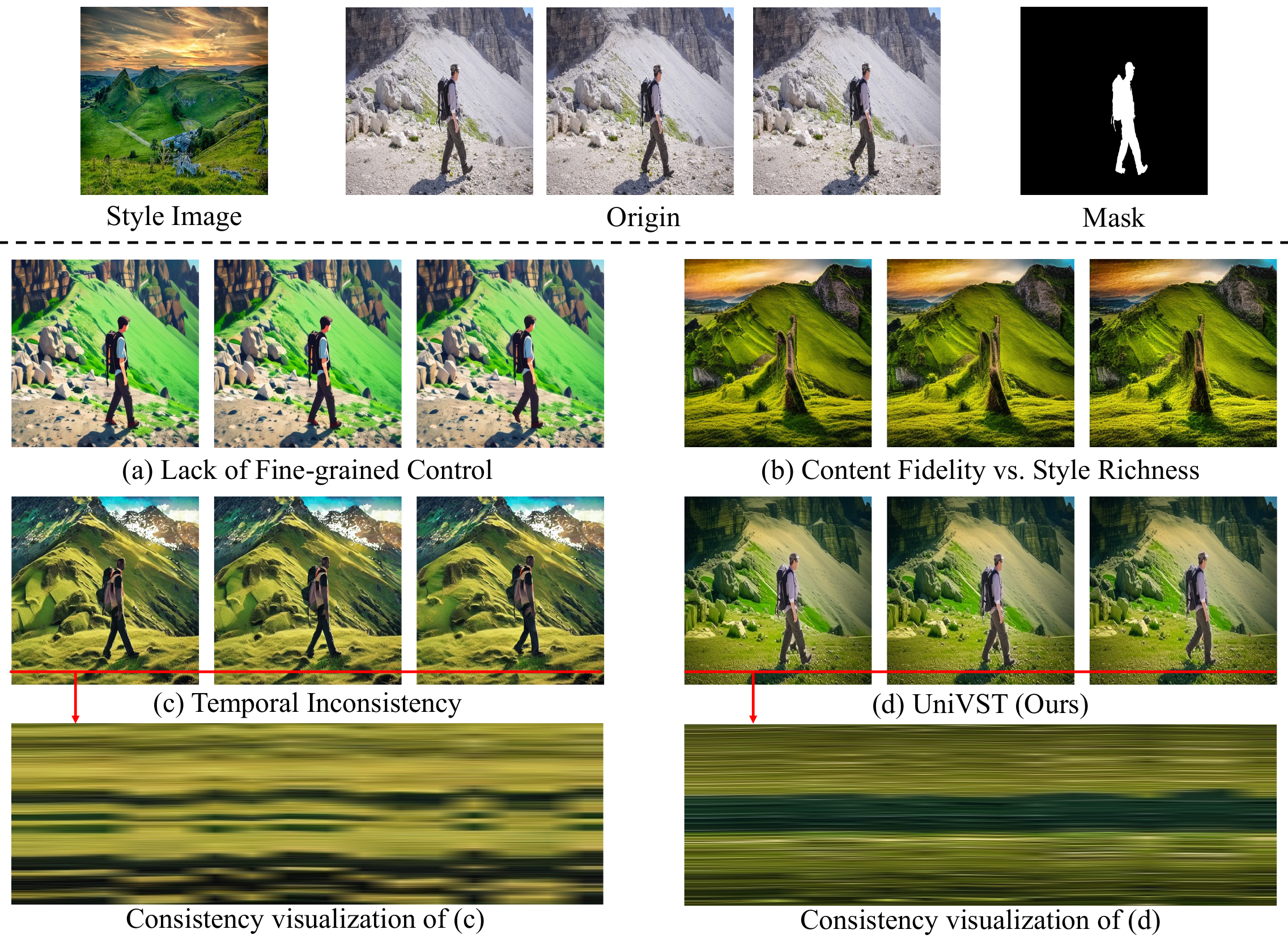} 
\caption{
Existing methods suffer from (a) lack of fine-grained control, (b) imbalance between content fidelity and style richness, and (c) temporal inconsistency, whereas our UniVST effectively overcomes these challenges.
The last row visualizes the coherence across frames using MimicMotion's Y-T slices~\cite{mimic_motion} for both BIVDiff~\cite{BIVDiff} and our UniVST.}
\vspace{-1.0em}
\label{fig:first_fig}
\end{figure}

Video stylization, a sub-area of video editing, is gaining popularity. It is about adding artistic styles to video content. Our review identifies two main video stylization methods.
The first method uses the image or text to direct the style. StyleCrafter~\cite{StyleCrafter} personalizes video style with an extra training adapter; VideoBooth~\cite{VideoBooth} uses the image for custom video creation prompts; while Diffutoon~\cite{Diffutoon} focuses on cartoon coloring from a text prompt.
The second method includes general video editing frameworks like AnyV2V~\cite{AnyV2V} and BIVDiff~\cite{BIVDiff} that can perform various tasks, including style transfer. These usually apply a consistent style across the whole video, which is good for thematic expression but sometimes lacks precision for specific uses. For example, as shown in Fig.\,\ref{fig:first_fig}, in film, scenes might be shot normally and later stylized to match different environments. This enables varied scene changes from one video take, which is also useful in advertising for efficiently applying local style changes to existing videos. Therefore, selectively styling parts of a video while keeping others unchanged is an important academic and commercial task.

We illustrate Fig.\,\ref{fig:first_fig} and pinpoint three major limitations of current methods:
\textit{1. Lack of Fine-grained Control}. While some existing approaches~\cite{Fate-Zero, Tune-A-Video} leverage localized text descriptions to preserve the overall style transfer and maintain the primary object's style, the control is usually too coarse. This lack of precision can lead to the model failing to understand the nuances, potentially causing unintended style transfer effects on objects not explicitly mentioned in the text.
\textit{2. Content Fidelity vs. Style Richness}.  In the realm of video stylization, there is a delicate balance to be struck between content fidelity and style richness. An overemphasis on style richness can result in blurring and altering the original layout of the video, whereas an excessive focus on content fidelity might yield a stylized outcome that is indistinguishable from the original.
\textit{3. Temporal Inconsistency}. Unlike image stylization, video stylization requires careful consideration of temporal coherence between frames. As in BIVDiff~\cite{BIVDiff} and AnyV2V~\cite{AnyV2V}, simply extending image stylization techniques to video often results in frame inconsistencies, causing various flickering and artifacts.

In this paper, we propose \textbf{UniVST}, which integrates three components into a single training-free framework for the task of localized\footnote{Here, ``localized'' indicates a particular part of the video.} video style transfer, providing a unified solution to the above three limitations.
The first step in achieving this is to obtain masks of the region specified by users, which allows for precise control during the subsequent stylization process.
However, unlike in the image domain~\cite{Repaint, MAG-Edit}, where masks can be readily obtained using additional models, the video domain incurs additional inference overhead and presents a more labor-intensive challenge due to the need for frame-by-frame mask generation.
Recognizing the high redundancy between video frames, we propose a \textit{Point-Matching Mask Propagation} strategy to efficiently extend the initial frame’s mask to subsequent frames. Leveraging feature maps from the inversion process~\cite{DDIM}, this strategy captures correlations in the latent space, eliminating the need for additional models.

By applying the DDIM inversion on both the original video and style image, we obtain the initial noise for each. Integrated with our point-matching mask propagation strategy, we can also obtain frame-by-frame masks for the object region specified by users.
Then, we establish new edited latents and construct a three-branch denoising architecture with content, edited, and style branches, for training-free \textit{AdaIN-Guided Localized Video Stylization}.
It first utilizes localized latent blending to fuse information from the content and edited branches, achieving localized style control during the style transfer process.
In addition, to enhance information interaction between the edited and style branches during denoising, we propose a training-free AdaIN-guided latent-shift and attention-shift mechanism.
Unlike traditional image stylization methods~\cite{StyleID}, It effectively mitigates detail loss during the stylization process, while also adeptly balancing content fidelity and style richness.

However, directly applying AdaIN-guided localized video stylization per-frame will introduce flickering and artifacts in final edited video. We empirically observe that this issue arises from a training-free stylization mechanism, which disrupts the temporal consistency of the edited latent.
Inspired by previous works~\cite{Ground-A-Video,Flatten}, we further employ optical flow to enhance temporal consistency in the latent space.
Specifically, we present a specialized \textit{Sliding-Window Consistent Smoothing} scheme during denoising, which employs optical flow for spatial alignment at the pixel level and refines predicted noise in the latent space.
As illustrated in Fig.\,\ref{fig:first_fig}, our smoothing scheme reduces flickering and artifacts while enhancing the temporal consistency in the final edited video.

Overall, our major contributions in this paper are as follows:
\begin{itemize}
\item We introduce, to the best of our knowledge, the first localized video stylization framework with diffusion models, featuring a novel point-matching mask propagation strategy.
\item We develop a training-free AdaIN-guided localized video stylization mechanism that operates at both the latent and attention levels. It effectively harmonizes content fidelity and style richness during the style transfer process.
\item We present a sliding-window consistent smoothing scheme based on optical flow, which adeptly upholds temporal consistency throughout the style transfer process.
\item Comprehensive experiments demonstrate that our framework outperforms several state-of-the-art methods across various qualitative and quantitative evaluation metrics.
\end{itemize}

\section{Related Work}

\noindent \textbf{Image Style Transfer.}
Image style transfer aims to endow an image with a specific style, with early research focusing on non-diffusion methods. RSCT~\cite{RSCT} calls for the separate transfer of foreground and background styles, combining them to ensure consistent styling. CAST~\cite{CAST} uses contrastive learning to extract style representations from image features, enhancing performance in arbitrary style transfer tasks. EFDM~\cite{EFDM} utilizes exact feature distribution matching to align complex feature distributions across diverse style domains. 
With the development of diffusion models, recent studies have increasingly explored their application in style transfer. DiffStyle~\cite{DiffStyler} employs LoRA~\cite{LoRA} to refine image prompts for stylization, while ArtBank~\cite{ArtBank} enhances text embeddings through fine-tuning image prompts. StyleID~\cite{StyleID} leverages key-value replacement coupled with an initial latent AdaIN~\cite{AdaIN} to preserve content fidelity. Z*~\cite{Z*} integrates content with diverse styles through an attention-based rearrangement mechanism. InstanceStylePlus~\cite{InstantStylePlus} utilizes adapters to safeguard content retention and enhance stylistic expression. Although these methods demonstrate superior performance, directly applying them to video tasks using frame-wise processing often fails to ensure semantic continuity.

\noindent \textbf{Video Style Transfer.}
Video style transfer prioritizes per-frame image style transfer while ensuring visual coherence. Similar to image style transfer, early research also concentrated on non-diffusion approaches. RT-NVST~\cite{real-time_nvst} uses feedforward networks for real-time video style transfer with a hybrid loss ensuring temporal consistency. RT-LVST~\cite{real-time_lvst} proposes a deep neural network for transferring artistic styles to specific video regions while preserving photorealism. These traditional non-diffusion models incur extra training costs and underperform diffusion models. 
Recent studies have increasingly employed diffusion models to accomplish video stylization. OCD~\cite{OCD} integrates object-centric sampling and merging, expediting the editing process. StyleCrafter~\cite{StyleCrafter} refines adapters using image prompts based on the T2V model. Style-A-Video~\cite{Style-A-Video} enhances stylistic alignment through strategic style guidance. Owing to the substantial computational demands of training, some studies have turned to training-free video stylization. Diffutoon~\cite{Diffutoon} pioneers primary and editing branches for multi-segment editing in video colorization. BIVDiff~\cite{BIVDiff} combines frame-by-frame image editing with temporal consistency modeling. AnyV2V~\cite{AnyV2V} follows a similar initial editing strategy, subsequently leveraging an I2V model~\cite{I2vgen-xl} to propagate changes across the video sequence. Although these diffusion-based methods exhibit strong performance, they struggle with fine-grained local editing while balancing content fidelity and style richness. In contrast, our proposed UniVST achieves fine-grained transfer while maintaining both content fidelity and style richness in a training-free manner.

\begin{figure*}[t]
\centering
\includegraphics[width=0.95\textwidth]{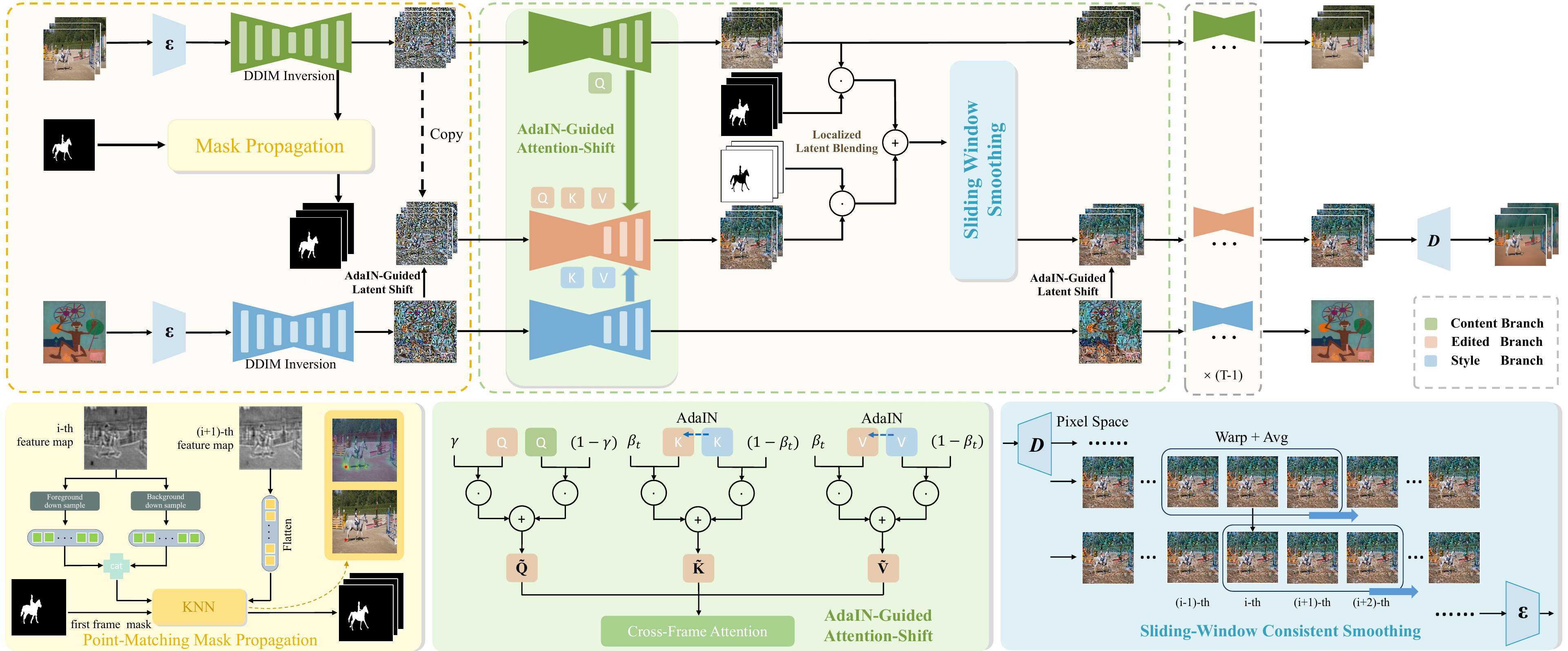}
\caption{Overall framework of our UniVST. 
It consists of three core components.
(1) \textit{Point-Matching Mask Propagation}, (2) \textit{AdaIN-Guided Localized Video Stylization} (Localized Latent Blending, AdaIN-Guided Latent-Shift, and Attention-Shift), and (3) \textit{Sliding-Window Consistent Smoothing.}}
\label{fig:overall_framework}
\vspace{-1em}
\end{figure*}

\section{Preliminary}
\textbf{Latent Diffusion Model.}
LDM~\cite{LDM} adds and removes noise in a low-dim space by using the encoder $\mathcal{E}$ and decoder $\mathcal{D}$. Given a Gaussian noise $Z_T$, DDIM denoising~\cite{rerender-a-video} can be formulated as:
\begin{equation}
Z_{t-1}=\sqrt{\alpha_{t-1}} \underbrace{{Z}_{t \rightarrow 0}}_{\text {predicted } Z_{0}}+\underbrace{\sqrt{1-\alpha_{t-1}} \epsilon_{\theta}\left(Z_{t}, t, C\right)}_{\text {direction pointing to } Z_{t-1}},
\label{eq:1}
\end{equation}
in which ${Z}_{t \rightarrow 0}$ is an approximate estimation of $Z_0$ at time step $t$:
\begin{equation}
{Z}_{t \rightarrow 0}=\left(Z_{t}-\sqrt{1-{\alpha}_{t}} \epsilon_{\theta}\left(Z_{t}, t, C\right)\right) / \sqrt{\alpha_{t}},
\label{eq:2}
\end{equation}
where $\alpha_t$ is predefined parameter and $C=\phi$ for editing integrity.

\noindent \textbf{DDIM Inversion.}
DDIM~\cite{DDIM} transforms noise $Z_{T}$ into $Z_{0}$. 
Given reversible ODE~\cite{ODE}, its reverse process can be described by:
\begin{equation}
{Z}_{t+1}={A}_{t} {Z}_{t} + {B}_{t} 
\epsilon_{\theta}\left({Z}_{t}, t, \phi \right),
\label{eq:3}
\end{equation}
where $A_{t}$ and $B_{t}$ are the functions of time step $t$. During editing, the initial noise $Z_{T}$, which closely aligns with the original video distribution $Z_{0}$, can be accurately derived using this equation.

\noindent \textbf{3D UNet Extension.} 
Following previous works~\cite{VDM, Flatten}, standard LDM will be extended into a 3D U-Net for video tasks. The $3\times3$ convolutional kernels in convolutional blocks are expanded to $1\times3\times3$. To enable frame parallelization, the intermediate features are reshaped from $\mathbb{R}^{b \times c \times f \times h \times w}$ to $\mathbb{R}^{(bf) \times c \times h \times w}$ before self-attention and cross-attention.
To enhance frame interaction, we follow T2V-Zero~\cite{T2V-Zero} to modify the self-attention into cross-frame attention:
\begin{equation}
\begin{array}{c}
    Q^i = W^{query} \cdot \mathcal{F}^{i}, \quad
    K^i = W^{key} \cdot [\mathcal{F}^{{1}}, \mathcal{F}^{{i-1}}], \\ 
    V^i = W^{value} \cdot [\mathcal{F}^{{1}}, \mathcal{F}^{{i-1}}],
\end{array}
\label{eq:4}
\end{equation}
where $\mathcal{F}^{i}$ denotes the extracted feature of the $i$-th frame, while $W^{query}$, $W^{key}$, and $W^{value}$ represent the corresponding learnable mapping matrices, with $[\cdot,\cdot]$ indicating concatenation operation.

\section{Localized Video Style Transfer}
Given an original video with $N$ frames $\{I^i\}_{i=1}^N$ and a style image $I^s$, our objective is to transform $\{I^i\}_{i=1}^N$ into a new sequence $\{\mathcal{I}^i\}_{i=1}^N$, which incorporates the reference style while preserving the primary objects' original style.
To achieve this, we propose UniVST, a unified framework for training-free localized video style transfer. The overall framework is illustrated in Fig.\,\ref{fig:overall_framework}.
UniVST first applies DDIM inversion to the original video and style image to obtain their initial noise and integrates \textit{Point-Matching Mask Propagation} to generate masks for the object regions.
It then performs \textit{AdaIN-Guided Localized Video Stylization} with a three-branch architecture for information interaction.
Moreover, \textit{Sliding-Window Consistent Smoothing} is incorporated into the denoising process, enhancing the temporal consistency in the latent space.

\begin{figure}[!t]
\centering
\includegraphics[width=0.95\columnwidth]{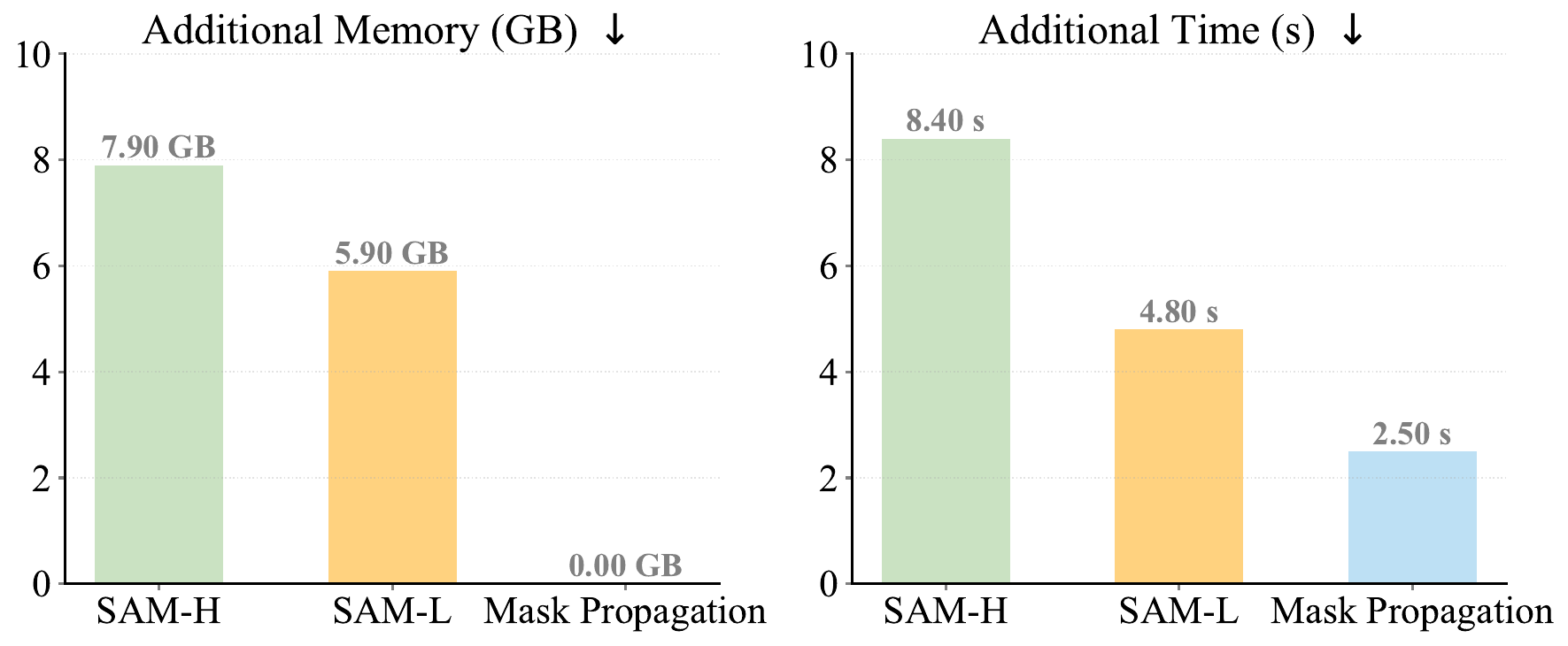}
\caption{The comparison between our mask propagation strategy and the traditional SAM model in terms of memory usage and inference time.
}
\label{fig:maskpropagation_intro}
\vspace{-1em}
\end{figure}

\subsection{Point-Matching Mask Propagation}
\begin{quote}
    \textit{``A towering building begins with a solid foundation.''}
\end{quote}
As the proverb clearly illustrates, the foundation is crucial for any significant endeavor. In the context of the localized video style transfer, the first essential step is to obtain precise and accurate masks for the region of interest specified by the users. These masks essentially form the cornerstone for enabling fine-grained control over specific regions during the subsequent stylization process.

Compared to image style transfer, the video domain presents a unique challenge: obtaining mask information for each frame. One potential solution is to employ segmentation models, such as SAM~\cite{SAM}, which can perform frame-by-frame segmentation based on the specified object. However, as shown in Fig.\,\ref{fig:maskpropagation_intro}, this approach has notable drawbacks. Specifically, employing SAM-H introduces an additional memory overhead of $7.9$ GB and increases the inference time, making it unfavorable for our UniVST.

Given the substantial redundancy between consecutive video frames, we propose leveraging this redundancy to efficiently propagate the initial frame's mask across all subsequent frames, thereby eliminating the need for additional segmentation models. This approach not only reduces memory usage but also accelerates the overall processing time. Thus, we introduce our point-matching mask propagation strategy, which is outlined in Algorithm\,\ref{Algorithm:1}.

\begin{algorithm}[!ht]
\scriptsize
\caption{Efficient Point-Matching Mask Propagation}
\begin{algorithmic}[1]
\STATE \textbf{Input:} $\mathcal{M}_1$ \hfill{The mask for the first frame.}
\STATE \textbf{Input:} $F_{t_0}$ \hfill{The feature map from last up-sampling block.}
\STATE \textbf{Input:} $r$ \hfill{The sampling rates.}
\STATE \textbf{Input:} $k$ \hfill{The $k$-nearest points.}
\STATE \textbf{Input:} $n$ \hfill{The number of reference frames.}
\STATE \textbf{Output:} $\{\mathcal{M}^i\}_{i=1}^N$ \hfill{The masks for all frames.}
\STATE \textbf{Initialize:} PreviousFeatures $\gets [\,]$ \hfill{Initialize empty list for previous features.}
\STATE \textbf{Initialize:} PreviousMasks $\gets [\,]$ \hfill{Initialize empty list for previous masks.}
\STATE \textbf{Initialize:} FirstFeature $\gets [\,]$ \hfill{Initialize empty list for first feature.}

\FOR{$i = 1, \dots, N$}
    \STATE $\text{ForeIndices} \gets \text{Where}(M_{i-1} = 1)$
    \STATE $\text{ForeIndices} \gets \text{Where}(M_{i-1} = 0)$
    \STATE $\text{ForeIndices} \gets \operatorname{RandomSample}(\text{ForeIndices}, r \cdot \frac{\lvert \text{ForeIndices} \rvert}{\lvert \text{ForeIndices} \rvert + \lvert \text{BackIndices} \rvert})$
    \STATE $\text{BackIndices} \gets \operatorname{RandomSample}(\text{BackIndices}, r \cdot \frac{\lvert \text{BackIndices} \rvert}{\lvert \text{ForeIndices} \rvert + \lvert \text{BackIndices} \rvert})$
    \STATE $\text{CurrentIndices} \gets \operatorname{Concat}(\text{ForeIndices}, \text{BackIndices})$
    \STATE $\text{CurrentFeature} \gets F_{t_0}^i[\text{CurrentIndices}]$

    \IF{$i = 1$}
        \STATE \text{FirstFeature.append}($\text{CurrentFeature}$) \hfill{Store the first frame.}
        \STATE $\mathcal{M}^i \gets \mathcal{M}^1$
    \ELSE
        \IF{$\lvert \text{PreviousFeatures} \rvert \geq n$} 
            \STATE \text{PreviousFeatures.pop}(0) \hfill{Remove the oldest frame feature.}
            \STATE \text{PreviousMasks.pop}(0) \hfill{Remove the oldest frame mask.}
        \ENDIF
        \STATE $f \gets \operatorname{Concat}(\text{PreviousFeatures}, \text{FirstFeature})$
        \STATE $\mathcal{M}^i \gets \text{KNN}(f, F_{t_0}^{i}, \operatorname{Concat}(\text{PreviousMasks}, \mathcal{M}^1), k)$
        \STATE \text{PreviousFeatures.append}($\text{CurrentFeature}$) \hfill{Store the current frame.}
        \STATE \text{PreviousMasks.append}($\mathcal{M}_i$) \hfill{Store the current mask.}
    \ENDIF
\ENDFOR

\STATE \textbf{Return:} $\{\mathcal{M}^i\}_{i=1}^N$
\end{algorithmic}
\label{Algorithm:1}
\end{algorithm}

\textbf{Point-Matching.}
Inspired by DIFT~\cite{DIFT}, the three up-sampling blocks in the U-Net are replete with rich semantic information. Therefore, during the DDIM inversion in the original video, we can retain the feature map $\{F_{t_0}^i\}_{i=1}^N$ from the last up-sampling block at a given step $t_0$, and use them to calculate the pixel correspondences across frames.
Specifically, to pinpoint the pixel in the $j$-th frame that most closely resembles a given point $p_i$ in the $i$-th frame, we calculate the cosine similarity between the corresponding point $p^{i}$ in the feature map $F_{t_0}^i$ and all points $p^{j}$ in the feature map $F_{t_0}^j$:
\begin{equation}
p^{j}={\underset{p^{j} \in F_{t_0}^j}{\arg\min}} \ \operatorname{CosSim}\big(p^{i}, p^{j}\big), \quad p^i \in F_{t_0}^i.
\label{eq:6}
\end{equation}
In which we can extend the initial frame’s mask across subsequent frames by the mask propagation strategy introduced below.

\textbf{Mask Propagation.}
Given an initial object mask $\mathcal{M}^{1}$ specified by the users, we use $k$-NN~\cite{KNN} to propagate this mask to subsequent frames. For each point in the $i$-th frame, we identify the $k$ most similar points in the $(i-1)$-th frame based on the point-matching principle described above. If the majority of these $k$ points are in the foreground, the corresponding point in the $i$-th frame is classified as the foreground; otherwise, it is classified as the background. In this way, we can obtain the masks $\{\mathcal{M}^{i}\}_{i=1}^{N}$ for all frames.

\begin{figure}[t]
\centering
\includegraphics[width=0.95\columnwidth]{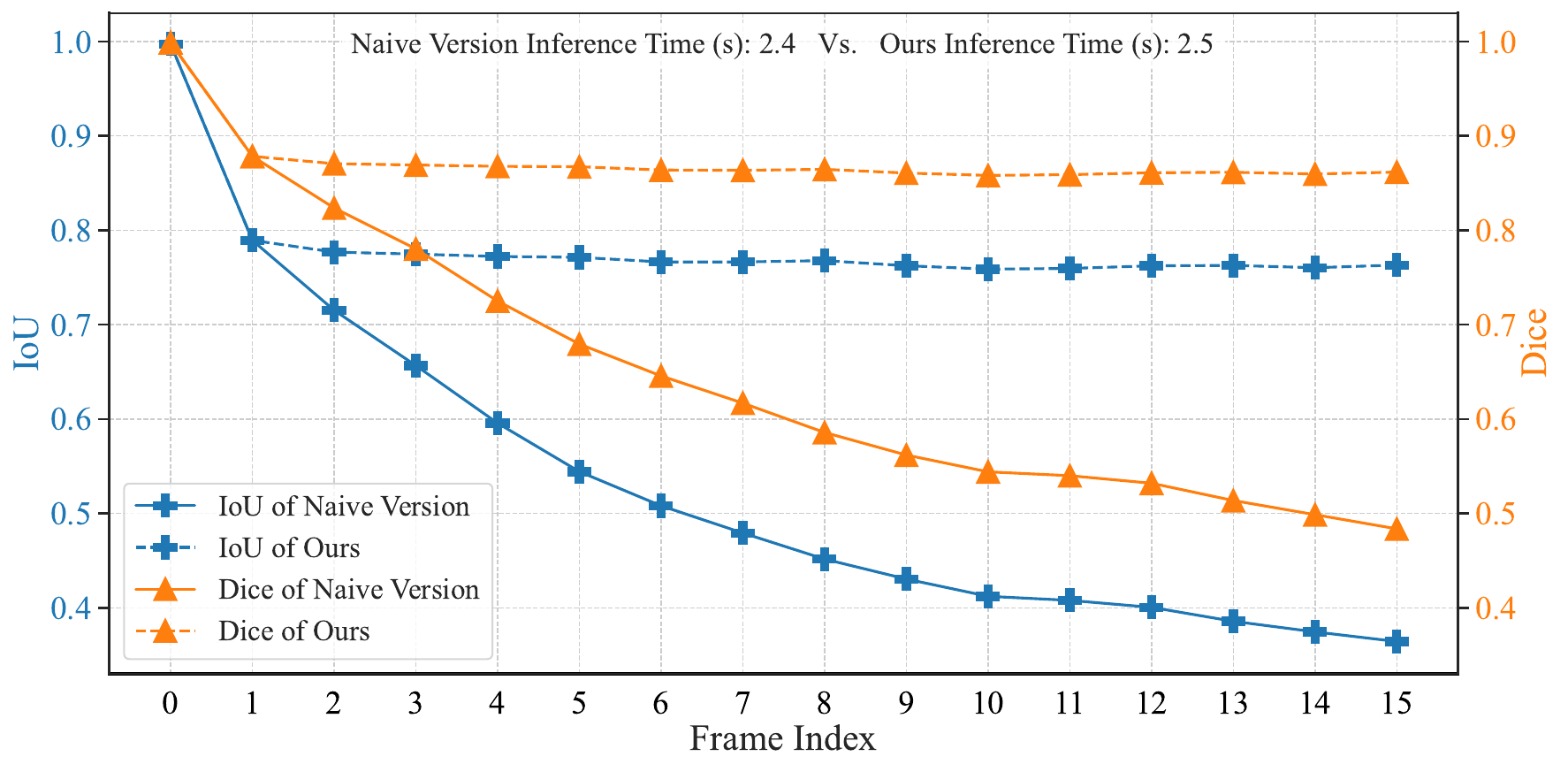}
\caption{
Comparison of accuracy and inference time under different mask propagation strategies. The accuracy is measured by relevant segmentation metrics: the intersection over union (IoU) and dice coefficients (Dice), which assess the overlap between predicted and ground truth regions. Our strategy effectively balances inference time and accuracy.
}
\label{fig:comparision of maskpropagation}
\vspace{-1em}
\end{figure}

However, as shown in Fig.\,\ref{fig:comparision of maskpropagation}, using a naive autoregressive approach for mask propagation will result in error accumulation, which degrades the accuracy of subsequent frames. To mitigate this issue, we design the anchor frames mechanism that incorporates information from the first frame as well as the previous $n$ frames.
On the other hand, although the introduction of the anchor frames mechanism improves the propagation accuracy of subsequent frames, it introduces additional similarity calculations, potentially reducing efficiency. To address this issue, we apply random down-sampling at a rate of $r$ to the foreground and background of the anchor frames, with the rate adjusted based on their respective proportions.
As illustrated in Fig.\,\ref{fig:comparision of maskpropagation}, the combination of anchor frames and random down-sampling strategies effectively preserves mask propagation accuracy while improving efficiency. 
More comprehensive analysis will be provided in Sec.\,\ref{sec:ablation_mask_propagation}.

\subsection{AdaIN-Guided Localized Video Stylization}
To achieve localized video stylization, we first obtain the initial latent of the original video $\{I^i\}_{i=1}^N$ and the style image $I^s$ using the inversion technique~\cite{easy_inv}, denoted as $Z_{t}$ and $Z_{t}^{s}$ respectively, for $t = 1 \rightarrow T$. Integrated with our point-matching mask propagation strategy, we can also obtain the mask frame-by-frame efficiently.
We then initialize the edited latent $\mathcal{Z}_{T} = Z_{T}$ and introduce a three-branch architecture with content, edited, and style branches, which perform the subsequent denoising process in parallel for $t = T \to 1$.
These branches will interact through \textit{localized latent blending}, \textit{AdaIN-guided latent-shift}, and \textit{AdaIN-guided attention-shift}, enabling precise control in subsequent video stylization.

\textbf{Localized Latent Blending.} 
To ensure that the style within the user-specified region is preserved throughout the denoising process, we utilize the masks $\{\mathcal{M}^{i}\}_{i=1}^{N}$, which is obtained from the previous mask propagation, as auxiliary constraints.
Specifically, during each denoising step, we blend the latent of content branch $Z_t^{i}$ and the latent of style branch $\mathcal{Z}_t^{i}$, using the mask $\mathcal{M}^{i}$ of the $i$-th frame:
\begin{equation}
\mathcal{Z}_{t}^{i} = \mathcal{M}^{i} \cdot Z_{t}^{i} + (1 - \mathcal{M}^{i}) \cdot \mathcal{Z}_{t}^{i}, \quad i = 1, ..., N
\end{equation}
This blending operation effectively ensures that the style within the user-specified region is seamlessly retained while allowing the rest of the frame to be fully stylized according to the edited branch.

\textbf{AdaIN-Guided Latent-Shift.}
AdaIN~\cite{AdaIN} is primarily used in image processing~\cite{StyleID}, where it adjusts the mean and standard deviation of an image to control its color style.
Inspired by this, we apply the AdaIN technique to integrate the color style of the reference image into the edited video.
Specifically, within the interval $t \in [\tau_0, \tau_1]$, we apply AdaIN operation to the latent $\mathcal{Z}_{t}$ of the editing branch, shifting its mean and standard deviation to match those in the latent $Z_{t}^{s}$ of the style branch, as follows:
\begin{equation}
\begin{aligned}
    \tilde{\mathcal{Z}}_{t} &= \text{AdaIN}(\mathcal{Z}_{t}, Z_{t}^{s}),
\end{aligned}
\label{eq:7}
\end{equation}
\begin{equation}
\text{AdaIN}(\mathcal{Z}_{t}, Z_{t}^{s})=
\sigma\left(Z_{t}^{s}\right)\left(\frac{\mathcal{Z}_{t}-\mu\left(\mathcal{Z}_{t}\right)}{\sigma\left(\mathcal{Z}_{t}\right)}\right)+\mu\left(Z_{t}^{s}\right),
\label{eq:8}
\end{equation}
where $\mu(\cdot)$ and $\sigma(\cdot)$ denote respective mean and standard deviation.

\begin{figure}[t]
\centering
\includegraphics[width=0.95\columnwidth]{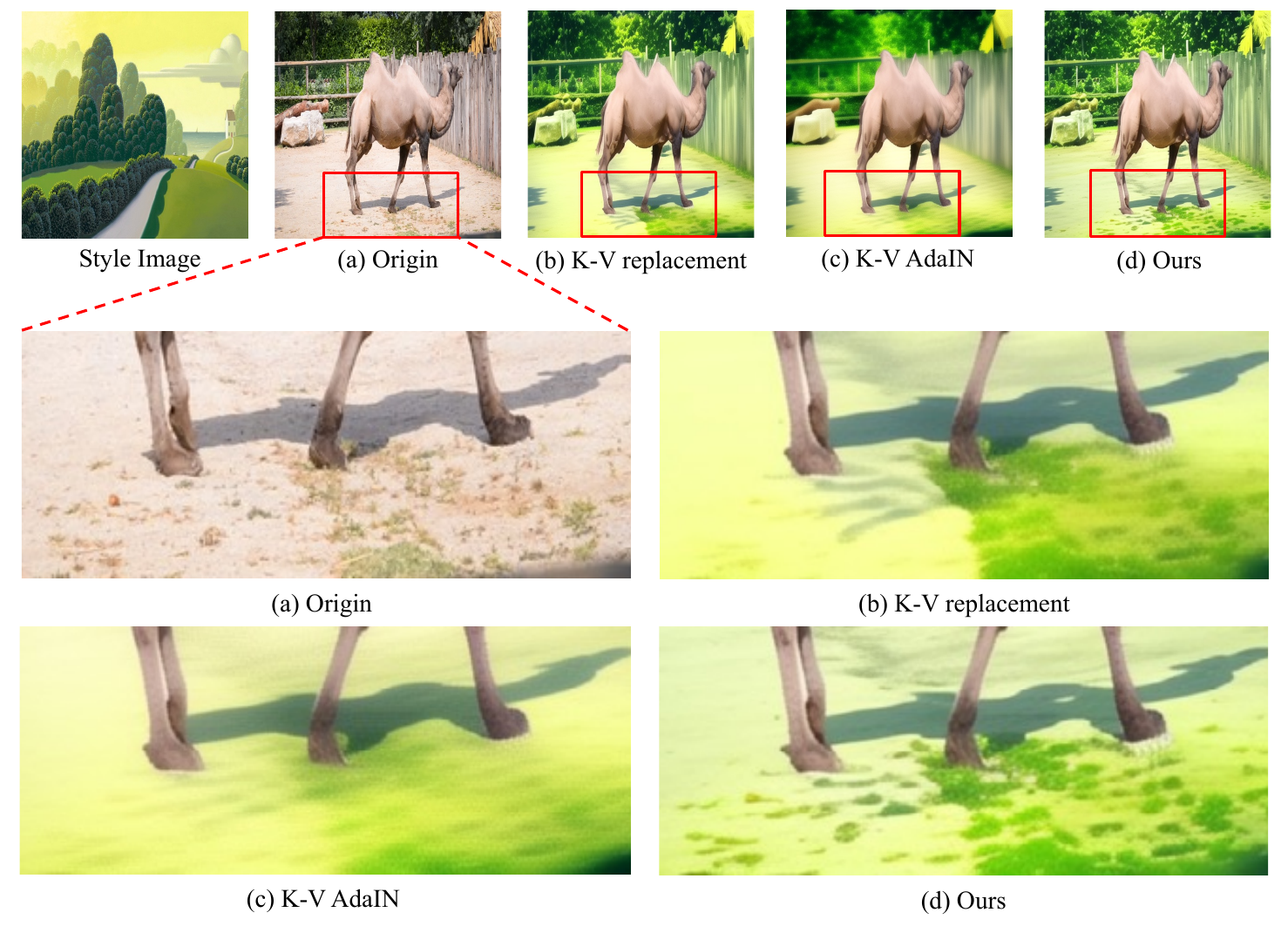}
\caption{Comparison of training-free stylization methods: (a) original video frame, (b) key-value replacement~\cite{StyleID}, (c) key-value AdaIN, and (d) AdaIN-guided attention-shift. Our approach mitigates local detail loss.
}
\label{fig:attention-shift}
\end{figure}

\textbf{AdaIN-Guided Attention-Shift.} 
In addition to transferring color tone, we also need to incorporate the overall style of the reference image into the edited video.
The previous work StyleID~\cite{StyleID} achieves style transfer by modifying the standard self-attention process, which is referred to as K-V replacement. Specifically, the query $\mathcal{Q}_t$ in the editing branch attends to the key $K_{t}^{s}$ and value $V_{t}^{s}$ of the style branch instead of its own. Furthermore, to eliminate content damage caused by direct replacement, \cite{StyleID} blends the original video's query $Q_t$ with the edited query $\mathcal{Q}_t$.
Finally, the resulting $\tilde{\mathcal{Q}}_{t}$, $\tilde{\mathcal{K}}_{t}$, and $\tilde{\mathcal{V}}_{t}$ undergo attention computation to produce the output $\mathcal{O}$. This entire process can be described as follows:
\begin{align}
\begin{split}
\tilde{\mathcal{Q}}_{t} &= \gamma \cdot \mathcal{Q}_{t} + (1-\gamma) \cdot Q_{t}, \\&
\tilde{\mathcal{K}}_{t}=K_{t}^{s}, \; \tilde{\mathcal{V}}_{t}=V_{t}^{s}, \\&
\mathcal{O} = \operatorname{Attention}(\tilde{\mathcal{Q}}_{t}, \tilde{\mathcal{K}}_{t}, \tilde{\mathcal{V}}_{t}),
\label{eq:9}
\end{split}
\end{align}
where $\gamma$ controls the balance between content and style. However, as observed in Fig.\,\ref{fig:attention-shift}(b), K-V replacement still results in slight local artifacts in the stylized videos. To further reduce the local artifacts caused by direct replacement, we apply AdaIN to $\mathcal{K}_t$ and $\mathcal{V}_t$ in the editing branch, shifting their mean and standard deviation to align with $K^s$ and $V^s$ in the style branch, resulting $\tilde{\mathcal{K}}_{t}$ and $\tilde{\mathcal{V}}_{t}$ as follows: 
\begin{equation}
\tilde{\mathcal{K}}_{t}=\text{AdaIN}(\mathcal{K}_{t}, K_{t}^{s}), \quad \tilde{\mathcal{V}}_{t}=\text{AdaIN}(\mathcal{V}_{t}, V_{t}^{s}).
\label{eq:10}
\end{equation}

Though the K-V AdaIN alleviates the artifact phenomenon, as indicated in Fig.\,\ref{fig:attention-shift}(c), it introduces new local detail blurring.
Upon analysis, we find that early denoising stages have less semantic information, making them suitable for K-V AdaIN, while later stages, containing more semantic information, are suited for key-value replacement.
In this way, within the interval $t \in [\tau_2, \tau_3]$, we gradually decrease the impact of the K-V AdaIN by combining the K-V replacements with a decreasing weight $\beta_t$ as follows:
\begin{equation}
\begin{aligned}
\tilde{\mathcal{K}}_{t} &= \beta_t \cdot \text{AdaIN}(\mathcal{K}_{t}, K_{t}^{s}) + (1 - \beta_t) \cdot K_{t}^{s}, \\
\tilde{\mathcal{V}}_{t} &= \beta_t \cdot \text{AdaIN}(\mathcal{V}_{t}, V_{t}^{s}) + (1 - \beta_t) \cdot V_{t}^{s}, \\
\beta_t &= \frac{\beta_{\tau_3} - \beta_{\tau_2}}{\tau_3 - \tau_2} \cdot (t - \tau_2) + \beta_{\tau_2},
\end{aligned}
\label{eq:11}
\end{equation}
in which $\beta_{\tau_2}$ and $\beta_{\tau_3}$ are the predefined weighting factors.

By dynamically combining K-V AdaIN with K-V replacement in Eq.\,(\ref{eq:11}) and query blending in Eq.\,(\ref{eq:9}), as shown in Fig.\,\ref{fig:attention-shift}(d), the stylization results achieve an effective balance between content fidelity and style richness.
Further quantitative comparison with other stylization methods will typically be presented in Sec.\,\ref{sec:ablation_localized_video_stylization}.

\subsection{Sliding-Window Consistent Smoothing}
Flickering and artifacts in the final edited video occur when AdaIN-guided localized video stylization is applied frame-by-frame. We empirically observe that this issue is caused by the training-free stylization mechanism, which disrupts the temporal consistency of the edited latent.
Prior works like MV-Diffusion~\cite{MV-Diffusion} leverage optical flow~\cite{Flatten, Ground-A-Video, MV-Diffusion} to model trajectory and motion patterns, thereby enhancing temporal consistency in video generation. While this approach is also feasible for the stylization process, it requires retraining modules, which conflicts with our training-free design philosophy.
By contrast, UniVST incorporates optical flow with specialized \textit{sliding-window consistent smoothing} to improve inter-frame coherence during denoising in a training-free manner.

\textbf{Optical Flow Warping.} 
Optical flow estimation is a pivotal technique widely employed in object tracking~\cite{MOT, OpticalTracking}. It estimates pixel-wise horizontal and vertical motion velocities by analyzing pixel variations across frames. Building on this, we utilize the optical flow map to align semantic content across frames, enhancing the coherence of the edited video.
Specifically, given two images, $A$ and $B$, we first employ RAFT~\cite{RAFT} to estimate the optical flow map from $A$ to $B$. This map is then used to transform $B$, aligning it with $A$ to ensure precise spatial correspondence. In cases of occlusion, bidirectional optical flow can identify occluded regions and recover them with relevant content from $A$. 
The above process, termed $\operatorname{Warp}(A, B)$, is essential for the following discussions.

\textbf{Sliding-Window Smoothing.} 
Rather than relying on simple post‐processing, we integrate the Warp operation into the denoising process to achieve local smoothing.
At time step $t$, we calculate the latent representation $\mathcal{Z}_{t \rightarrow 0}$ from the predicted noise $\varepsilon_{t}$ according to Eq.\,(\ref{eq:2}).
$\mathcal{Z}_{t \rightarrow 0}$ is then decoded by the decoder $\mathcal{D}$ into a pixel representation $\{P_{t}^i\}_{i=1}^N$.
For each decoded frame $P_t^i$, we apply a window of size $2m$ to warp neighboring frames onto it, followed by averaging to smooth the original pixel of $P_t^i$, as follows:
\begin{equation}
\bar{P}_{t}^{i} \leftarrow \frac{1}{2m+1} \cdot \sum_{j=i-m}^{i+m} \operatorname{Warp}(P_{t}^{i}, P_{t}^{j}).
\label{eq:12}
\end{equation}

The smoothed frame from the current window is used in the next window's calculations, as shown in Fig.\,\ref{fig:overall_framework}. 
Next, the VAE's encoder $\mathcal{E}$ re-encodes the smoothed $\{\bar{P}_{t}^i\}_{i=1}^N$ into latent representations $\bar{\mathcal{Z}}_{t \rightarrow 0}$, which is a refined version of $\mathcal{Z}_{t \rightarrow 0}$. 
Subsequently, the predicted noise $\varepsilon_{t}$ is adjusted using the inversion of Eq.\,(\ref{eq:2}):
\begin{equation}
\bar{\varepsilon}_{t} \leftarrow \frac{\mathcal{Z}_{t}-\sqrt{\alpha_{t}} \bar{\mathcal{Z}}_{t \rightarrow 0}}{\sqrt{1-\alpha_{t}}}.
\label{eq:13}
\end{equation}

Finally, the latent representation is updated according to the DDIM schedule in Eq.\,(\ref{eq:1}), enhancing coherence across frames:
\begin{equation}
\mathcal{Z}_{t-1} \leftarrow \sqrt{\alpha_{t-1}} \bar{\mathcal{Z}}_{t \rightarrow 0}+\sqrt{1-\alpha_{t-1}} \bar{\varepsilon}_{t}.
\label{eq:14}
\end{equation}

The above sliding-window consistent smoothing scheme is performed only within the interval $t \in [\tau_4, \tau_5]$.
Moreover, due to the challenges of optical flow estimation for fast-moving objects, we avoid global smoothing for long-range spatial alignment, employing instead a local sliding-window for short-range spatial alignment.

\section{Experimentation}
\subsection{Experimental Settings}
The main experiments of our proposed UniVST are based on the SD-v1.5~\cite{LDM}.
The hyper-parameters are configured as follows: $\tau_0 = 0.1T$, $\tau_1 = 0.2T$, $\tau_2 = 0.4T$, $\tau_3 = 1.0T$, $\tau_4 = 0.5T$, $\tau_5 = 0.6T$, $t_0 = 0.4T$, with the total time period $T$ set to 50. Additionally, we set $\gamma = 0.35$, $\beta_{\tau_2} = 0.1$, $\beta_{\tau_3} = 0.9$, $r = 0.3$, $k = 15$, $m = 2$, and $n = 9$. 
Original videos are resized to $512 \times 512$ pixels and processed in batches of $16$ frames. We use the RAFT model~\cite{RAFT} for optical flow estimation during the denoising process.
All main experiments can be executed on a single RTX 3090 GPU in a training-free manner.
To further validate the generalization and scalability of the proposed UniVST, we extend it to four additional representative backbones, including SD-v2.1~\cite{LDM}, SD-v3.0~\cite{SD3}, SD-v3.5~\cite{SD3}, and AnimateDiff-v2~\cite{AnimateDiff}, with both quantitative and qualitative comparisons.
Given that SD-v3.0 and SD-v3.5 are rectified-flow models, we employ RF-Solver~\cite{RF-Solver}, which is specifically designed for rectified flows, to replace the original DDIM inversion.

\textbf{Datasets Details.}
Building upon the foundational work referenced in~\cite{StyleID, Style-A-Video}, we have meticulously selected style images from two renowned databases: WikiArt~\cite{wikiart} and Laion-Aesthetics-6.5+~\cite{laion}, each contributing a collection of 20 distinct images. 
For the content videos, we have procured a dataset comprising 50 videos from the DAVIS2016~\cite{davis} and TGVE~\cite{tgve} datasets, which we combined and referred to as DAVTG. 
Through the integration of these datasets, we have successfully established two comprehensive subsets: DAVTG-WikiArt, which contains 1,000 curated pairs, and DAVTG-Laion, which also features 1,000 paired entries.

\begin{table*}[!t]
    \caption{Quantitative comparison on different datasets, where * denotes non-diffusion methods. 
    The \textbf{bold} and \underline{underline} indicate the best and second-best.}
    \centering
    \begin{tabular}{l ccc cc c c c} 
        \toprule
        \multirow{2}{*}{Methods} 
        & \multicolumn{3}{c}{Overall transfer performance} 
        & \multicolumn{2}{c}{Local style preservation} 
        & Consistency & Efficiency 
        & \multicolumn{1}{c}{Overall Quality} \\  
        \cmidrule(lr){2-4}  \cmidrule(lr){5-6}  \cmidrule(lr){7-7}  \cmidrule(lr){8-8} \cmidrule(lr){9-9}
        & ArtFID$\downarrow$ & FID$\downarrow$ & LPIPS$\downarrow$ 
        & SSIM$\uparrow$ & CLIP-I$\uparrow$ & CLIP-F$\uparrow$ 
        & Time(s)$\downarrow$ & VBench(\%)$\uparrow$ \\  
        \midrule
        \multicolumn{9}{c}{DAVTG-WikiArt} \\
        Diffutoon~\cite{Diffutoon} & 43.150 & 26.268 & \underline{0.582} & \underline{0.961} & \underline{0.954} & \textbf{0.980} & 44.263 & 72.251 \\  
        StyleCrafter~\cite{StyleCrafter} & 39.119 & \underline{21.008} & 0.777 & 0.928 & 0.895 & \underline{0.978} & 111.225 & 70.759 \\  
        AnyV2V~\cite{AnyV2V} & \textbf{33.774} & \textbf{19.104} & 0.680 & 0.938 & 0.914 & \underline{0.978} & 78.474 & \underline{72.791} \\  
        BIVDiff~\cite{BIVDiff} & 46.071 & 25.468 & 0.740 & 0.921 & 0.894 & 0.955 & 158.367 & 70.351 \\  
        EFDM$^*$~\cite{EFDM} & 37.407 & 22.350 & 0.602 & 0.936 & 0.935 & 0.963 & \textbf{5.969} & 71.649 \\  
        CAST$^*$~\cite{CAST} & 45.198 & 26.174 & 0.663 & 0.948 & 0.936 & 0.975 & \underline{7.348} & 70.429 \\ 
        \rowcolor{gray!25}
        UniVST (SD-v1.5) & \underline{37.152} & 26.134 & \textbf{0.369} & \textbf{0.986} & \textbf{0.991} & \textbf{0.980} & 162.745 & \textbf{74.014} \\  
        \midrule
        \multicolumn{9}{c}{DAVTG-Laion} \\
        Diffutoon~\cite{Diffutoon} & 38.610 & 23.700 & \underline{0.563} & \underline{0.959} & \underline{0.953} & \underline{0.979} & 44.263 & \underline{73.607} \\  
        StyleCrafter~\cite{StyleCrafter} & 43.161 & 23.246 & 0.780 & 0.926 & 0.897 & 0.978 & 111.225 & 71.379 \\  
        AnyV2V~\cite{AnyV2V} & 35.648 & \underline{20.019} & 0.696 & 0.936 & 0.918 & 0.975 & 78.474 & 71.927 \\  
        BIVDiff~\cite{BIVDiff} & 36.676 & 20.156 & 0.733 & 0.921 & 0.897 & 0.956 & 158.367 & 70.273 \\  
        EFDM$^*$~\cite{EFDM} & 35.673 & 21.549 & 0.582 & 0.933 & 0.931 & 0.964 & \textbf{5.969} & 70.628 \\  
        CAST$^*$~\cite{CAST} & \underline{34.828} & \textbf{19.974} & 0.661 & 0.947 & 0.936 & 0.968 & \underline{7.348} & 71.906 \\
        \rowcolor{gray!25}
        UniVST (SD-v1.5) & \textbf{30.636} & 21.146 & \textbf{0.383} & \textbf{0.986} & \textbf{0.990} & \textbf{0.981} & 162.745 & \textbf{74.495} \\  
        \bottomrule
    \end{tabular}
    \label{tab:comparison_all}
\end{table*}

\textbf{Baselines.}
We compare our UniVST with six advanced methods. These include two video stylization techniques: Diffutoon~\cite{Diffutoon} and StyleCrafter~\cite{StyleCrafter}; two video editing frameworks: AnyV2V~\cite{AnyV2V} and BIVDiff~\cite{BIVDiff}; and two traditional non-diffusion methods: EFDM~\cite{EFDM} and CAST~\cite{CAST}.
Since our specific task does not perfectly align with some of the existing methods, we have made the necessary adjustments to ensure a fair comparison.
In the case of Diffutoon~\cite{Diffutoon}, which is designed to process text inputs for coloring tasks and does not natively support the input of style images. To overcome this limitation, we leverage GPT-4o~\cite{GPT4} to generate descriptive captions for the style images.
For the style generation task with StyleCrafter~\cite{StyleCrafter}, which does not inherently support video editing, we employ inversion to obtain the initial noise from the original video, which is then utilized in the generation process.
For the unified video editing frameworks, AnyV2V~\cite{AnyV2V} and BIVDiff~\cite{BIVDiff}, we have integrated InstantStyle~\cite{InstantStyle} as the image style transfer model to perform the video style transfer task.
Regarding the traditional non-diffusion methods, EFDM~\cite{EFDM} and CAST~\cite{CAST}, which are well-aligned with our task, we have retained them in their original form without any modifications.

\begin{figure*}[!t]
\centering
\includegraphics[width=0.95\textwidth]{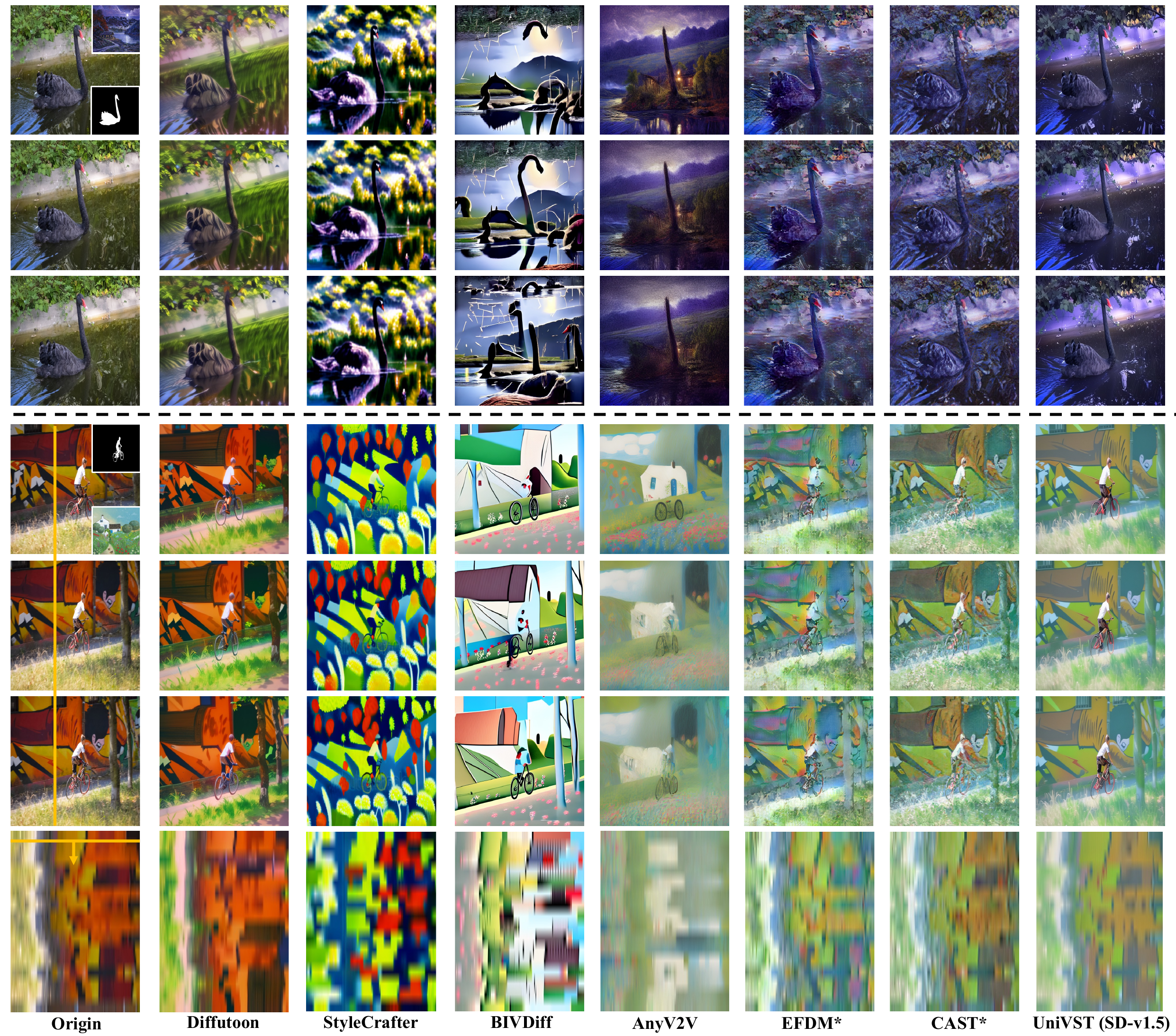}
\caption{
Qualitative comparison of our proposed UniVST with existing methods. Superscript * indicates non-diffusion methods.
}
\vspace{-0.5em}
\label{fig:qualitative_comparison}
\end{figure*}

\subsection{Quantitative Comparison}
\label{sec:quantitative_comparison}
To comprehensively assess the performance of different methods,
we evaluate their stylized videos across four aspects:
(1) \textit{Overall transfer performance.} Drawing inspiration from the methodologies outlined in~\cite{StyleID, Style-A-Video}, we assess the overall transfer performance using ArtFID~\cite{Artfid}, a comprehensive metric that combines LPIPS~\cite{LPIPS} for gauging content fidelity and FID~\cite{FID} for assessing style richness.
(2) \textit{Local style preservation.} To evaluate the preservation of local style, we utilize SSIM~\cite{SSIM} metric for the texture preservation, while CLIP-I score~\cite{clip} is utilized to assess the preservation of semantic content. During the evaluation, the background is neutralized to ensure accuracy.
(3) \textit{Temporal consistency.} In line with the approaches detailed in~\cite{Style-A-Video, StyleCrafter}, we measure the similarity between consecutive frames using the CLIP-F score~\cite{clip}, which serves as a measure of inter-frame coherence in the edited video.
(4) \textit{Inference efficiency.} To assess the inference efficiency, we calculate the aggregate time during inference, thereby assessing their performance in terms of computational speed.
(5) \textit{Overall quality.} Following the methodology introduced by VBench~\cite{VBench}, we comprehensively evaluate stylized videos in terms of subject consistency, background consistency, motion smoothness, dynamic degree, aesthetic quality, imaging quality, \emph{etc}., and report the averaged score to measure their overall quality.
Furthermore, we employ six established methods as baselines and benchmark them against our UniVST on two datasets: DAVTG-WikiArt and DAVTG-Laion.
Quantitative comparisons are comprehensively shown in Table\,\ref{tab:comparison_all}.

The results demonstrate that our UniVST outperforms all competing methods across three performance metrics. On both the DAVTG-WikiArt and DAVTG-Laion datasets, UniVST consistently surpasses existing methods in local style preservation and temporal consistency, showcasing its superiority in localized video style transfer tasks over existing methods.
Regarding overall transfer performance, UniVST achieves the best results on the DAVTG-Laion dataset and secures the second spot on DAVTG-WikiArt.
Although localized style transfer tasks may lead to a marginally higher FID score, slightly affecting overall performance compared to other methods, our UniVST still achieves a favorable balance between content fidelity and style richness.
Moreover, VBench evaluation further highlights the advantage of our UniVST, where it achieves the highest average score, demonstrating superior performance across multiple perceptual dimensions, including consistency, aesthetics, smoothness, artistic quality, and others.

\begin{figure*}[!b]
\centering
\includegraphics[width=0.95\linewidth]{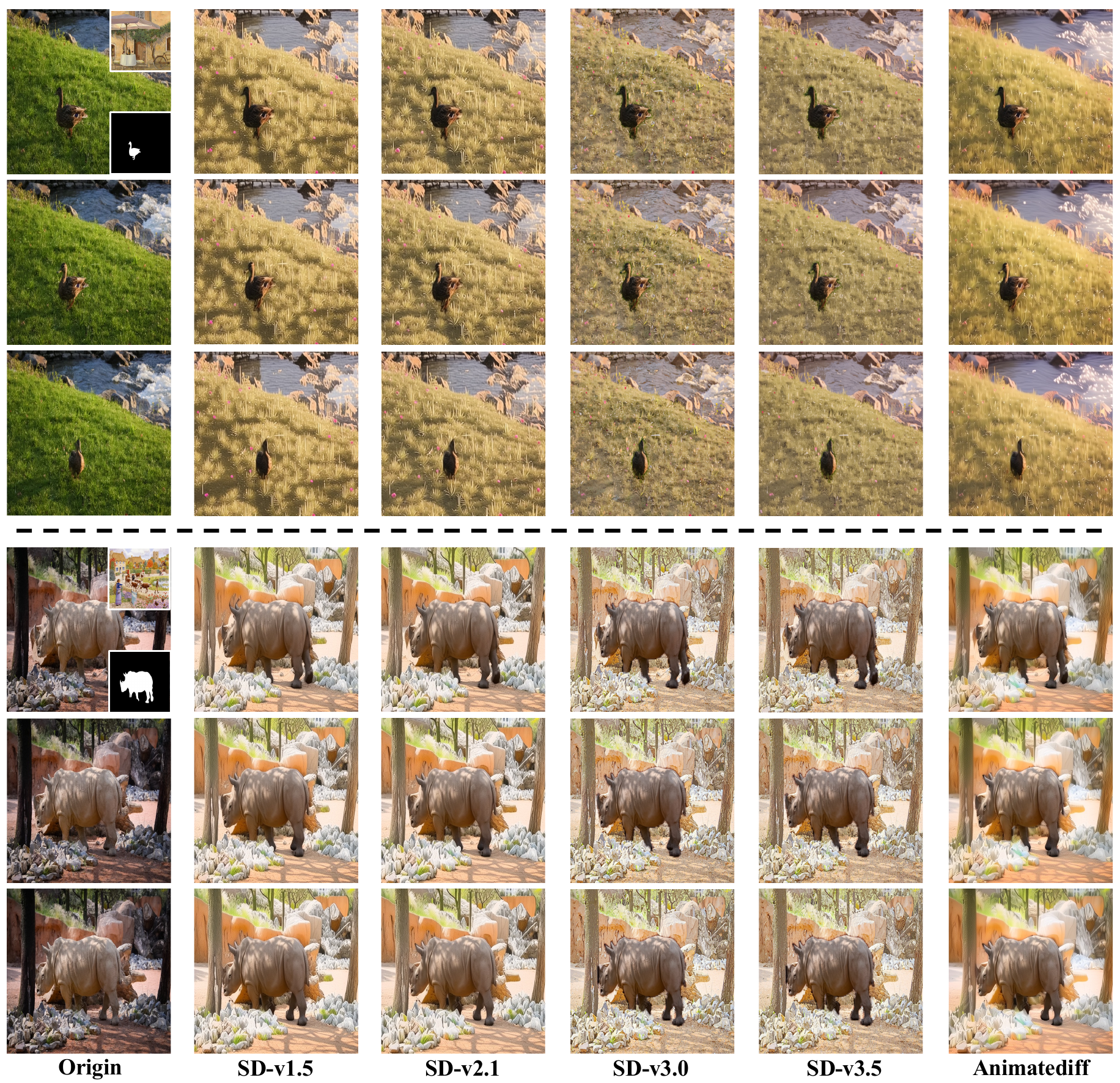}
\caption{
Additional visual results of UniVST on different backbones: SD-v1.5~\cite{LDM}, SD-v2.1~\cite{LDM}, SD-v3.0~\cite{SD3}, SD-v3.5~\cite{SD3} and Animatediff~\cite{AnimateDiff}.
}
\vspace{0.85em}
\label{fig:additioanl_visual_results}
\end{figure*}

\begin{table}[!t]
\centering
\caption{
Additional quantitative evaluation results of UniVST on different backbones on the DAVTG-Laion dataset.
}
\resizebox{\linewidth}{!}{
\begin{tabular}{lccccc}
\toprule
Metrics & SD-v1.5 & SD-v2.1 & SD-v3.0 & SD-v3.5 & Animatediff-v2 \\
\midrule
ArtFID$\downarrow$ & 30.636 & 30.261 & 28.720 & 28.550 & 30.490 \\
SSIM$\uparrow$   & 0.986 & 0.985 & 0.996 & 0.995 & 0.986 \\
CLIP-I$\uparrow$ & 0.990 & 0.989 & 0.997 & 0.997 & 0.990 \\
VBench(\%)$\uparrow$ & 74.495 & 74.645 & 75.622 & 75.683 & 74.575 \\
\bottomrule
\end{tabular}
}
\vspace{-1.0em}
\label{tab:additional_quantitative_results}
\end{table}

Our inference efficiency does not stand out significantly. This is because we opt not to adopt the temporal attention mechanisms in existing T2V frameworks~\cite{StyleCrafter, Style-A-Video}. Instead, we utilize the sliding-window consistent smoothing scheme, which requires dynamic optical flow estimation during the subsequent denoising process. While this approach improves temporal consistency, it also leads to increased computational overhead.
The detailed discussion on the inference efficiency of each module will be provided in Sec.\,\ref{sec:additional_analysis}.

To further evaluate the generalization capability of UniVST, we conduct additional quantitative experiments on the DAVTG-Laion dataset using four backbones: SD-v2.1~\cite{LDM}, SD-v3.0~\cite{SD3}, SD-v3.5~\cite{SD3}, and Animatediff-v2~\cite{AnimateDiff}. The corresponding results are shown in Table\,\ref{tab:additional_quantitative_results}.
On SD-v2.1 and Animatediff-v2, UniVST achieves ArtFID, SSIM, CLIP-I, and VBench scores comparable to those on SD-v1.5, indicating stable transfer performance, local style preservation, and overall quality. This highlights the strong generalization capability of UniVST.
When applied to the advanced rectified-flow models SD-v3.0 and SD-v3.5, UniVST surpasses its performance on SD-v1.5 across all metrics, demonstrating its scalability with increasingly powerful backbones.

\subsection{Qualitative Comparison}
For qualitative evaluation, we have juxtaposed the stylized videos with those of six baseline methods, as illustrated in Fig.\,\ref{fig:qualitative_comparison}.
Diffutoon~\cite{Diffutoon}, which is renowned for its expertise in cartoon coloring, exhibits limited responsiveness to a variety of image styles. This results in a final edited video that is notably monotonous, and heavily inclined towards a cartoonish anime aesthetic.
StyleCrafter~\cite{StyleCrafter}, tailored for style-based video generation, demonstrates a high sensitivity to stylistic cues. However, it encounters difficulties in preserving content fidelity, leading to substantial distortion of background details and character features during the transfer process.
BIVDiff~\cite{BIVDiff} falls short in both content fidelity and temporal consistency, similarly altering background information and character traits during the transfer.
AnyV2V~\cite{AnyV2V} excels in maintaining temporal consistency but faces challenges with transfer tasks involving a central object. It often over-transfers the subject's style, resulting in a loss of its distinctive attributes.
As for the traditional non-diffusion methods, EFDM~\cite{EFDM} and CAST~\cite{CAST}, they degrade fine details and induce color artifacts during the style transfer process.
On the contrary, our UniVST exhibits a remarkable ability to balance content fidelity and style richness, achieving a harmonious integration of both. It ensures that the key structural and motion details of the original videos are retained, while preserving the primary style of the object with great precision. 
Following MimicMotion~\cite{mimic_motion}, we use the slice to further visualize the temporal consistency of stylized videos across different methods, with the results shown in Fig.\,\ref{fig:qualitative_comparison}. The fluctuation trend of the slice obtained with our UniVST is most similar to that of the original video, demonstrating that UniVST achieves the smallest temporal consistency loss during the style transfer process.

\begin{table*}[!t]
    \caption{
    Ablation studies across different datasets. \textbf{Bold} indicate the best performance.
    }
    \centering
    \begin{tabular}{l ccc cc c c} 
        \toprule
        \multirow{2}{*}{Model Variants} 
        & \multicolumn{3}{c}{Overall Transfer Performance} 
        & \multicolumn{2}{c}{Local Style Preservation} 
        & Consistency 
        & \multicolumn{1}{c}{Overall Quality} \\  
        \cmidrule(lr){2-4}  \cmidrule(lr){5-6}  \cmidrule(lr){7-7} \cmidrule(lr){8-8}
        & ArtFID$\downarrow$ & FID$\downarrow$ & LPIPS$\downarrow$ 
        & SSIM$\uparrow$ & CLIP-I$\uparrow$ & CLIP-F$\uparrow$ 
        & VBench (\%)$\uparrow$ \\  
        \midrule
        \multicolumn{8}{c}{DAVTG-WikiArt} \\
        W/o all components (baseline) & 45.676 & 31.380 & 0.411 & 0.979 & 0.976 & 0.981 & 71.423 \\  
        W/o localized latent blending & 38.014 & \textbf{26.050} & 0.405 & 0.966 & 0.970 & 0.979 & 73.551 \\  
        W/o AdaIN-guided latent-shift & 37.649 & 26.909 & \textbf{0.349} & 0.985 & 0.990 & 0.980 & 73.655 \\  
        W/o AdaIN-guided attention-shift & 39.589 & 27.506 & 0.388 & 0.985 & 0.988 & \textbf{0.983} & 72.958 \\  
        W/o sliding-window consistent smoothing & 38.626 & 26.907 & 0.384 & 0.982 & 0.989 & 0.970 & 73.701 \\
        \rowcolor{gray!25}
        W/ all components (ours) & \textbf{37.152} & 26.134 & 0.369 & \textbf{0.986} & \textbf{0.991} & 0.980 & \textbf{74.014} \\  
        \midrule
        \multicolumn{8}{c}{DAVTG-Laion} \\
        W/o all components (baseline) & 38.981 & 26.594 & 0.413 & 0.979 & 0.976 & \textbf{0.981} & 71.571 \\  
        W/o localized latent blending & 30.848 & \textbf{20.974} & 0.404 & 0.961 & 0.967 & 0.979 & 73.888 \\  
        W/o AdaIN-guided latent-shift & 31.001 & 21.766 & \textbf{0.362} & 0.985 & 0.989 & 0.980 & 73.987 \\  
        W/o AdaIN-guided attention-shift & 32.534 & 22.623 & 0.377 & 0.985 & 0.989 & \textbf{0.981} & 73.395 \\  
        W/o sliding-window consistent smoothing & 30.834 & 21.249 & 0.386 & 0.981 & 0.988 & 0.971 & 74.044 \\
        \rowcolor{gray!25}
        W/ all components (ours) & \textbf{30.636} & 21.146 & 0.383 & \textbf{0.986} & \textbf{0.990} & \textbf{0.981} & \textbf{74.495} \\  
        \bottomrule
    \end{tabular}
    \label{tab:ablation_study_all}
\end{table*}

Moreover, we present additional visualizations of UniVST applied to four additional backbones, including SD-v2.1~\cite{LDM}, SD-v3.0~\cite{SD3}, SD-v3.5~\cite{SD3}, and Animatediff-v2~\cite{AnimateDiff}, to further highlight its generalization and scalability.
As presented in Fig.\,\ref{fig:additioanl_visual_results}, different backbones consistently generate similar stylization results, underscoring the stability of our framework.
It is also worth noting that UniVST also achieves stable visual performance under the state-of-the-art rectified-flow models such as SD-v3.0 and SD-v3.5, in practice requiring only the replacement of the inversion technique, thereby further demonstrating its scalability.

\begin{figure*}[!t]
\centering
\includegraphics[width=0.95\textwidth]{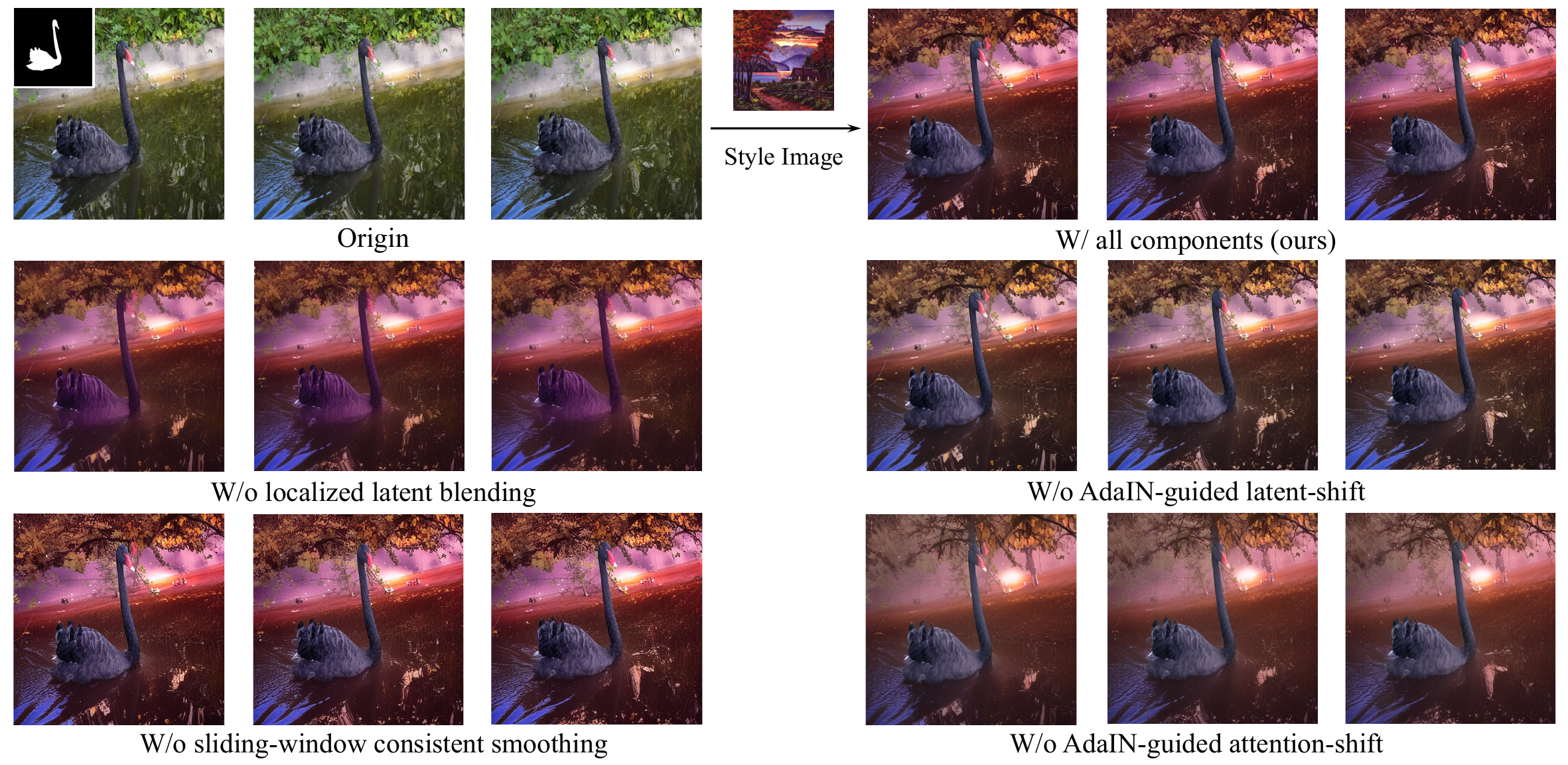}
\caption{Visual results for ablation studies. Various components in our framework have all played important roles.
}
\vspace{-1.5em}
\label{fig:abalation_study}
\end{figure*}

\subsection{Ablation Studies}
\label{sec:ablation_study}
In this section, we first validate the effectiveness of each core component in our proposed UniVST. Then, we conduct an in-depth analysis of hyper-parameter settings for the three modules: \textit{point-matching mask propagation}, \textit{AdaIN-guided localized video stylization}, and \textit{sliding-window consistency smoothing}.
All ablation experiments are conducted on SD-v1.5.

\subsubsection{Ablation Study on Core Components}
We have developed six distinct variants to assess the contribution of individual components in our proposed UniVST: (1) UniVST without all components (baseline), (2) UniVST without localized latent blending, (3) UniVST without AdaIN-guided latent-shift, (4) UniVST without AdaIN-guided attention-shift, (5) UniVST without sliding-window consistent smoothing and (6) UniVST with all components (ours). The quantitative outcomes are presented in Table\,\ref{tab:ablation_study_all}, while the qualitative results are displayed in Fig.\,\ref{fig:abalation_study}. 
Our localized latent blending strategy effectively preserves the foreground style during the stylization process, demonstrating its efficacy quantitatively and qualitatively, without the need for additional models like SAM~\cite{SAM}. The AdaIN-guided latent-shift and attention-shift strategies are essential for style transfer. Notably, omitting the attention-shift strategy leads to incomplete transfer and blurred local details, as shown in Fig.\,\ref{fig:abalation_study}. Moreover, our sliding-window consistency smoothing scheme enhances temporal consistency and effectively reduces video artifacts.
Overall, the coordinated effect of these components yields the lowest ArtFID~\cite{StyleID}, indicating optimal content–style balance; the highest SSIM~\cite{SSIM} and CLIP-I~\cite{clip}, reflecting effective local preservation; and the highest VBench score, showing superior overall quality across multiple dimensions, including consistency, aesthetics, art, and so on.

\begin{figure}[!t]
\centering
\includegraphics[width=0.95\linewidth]{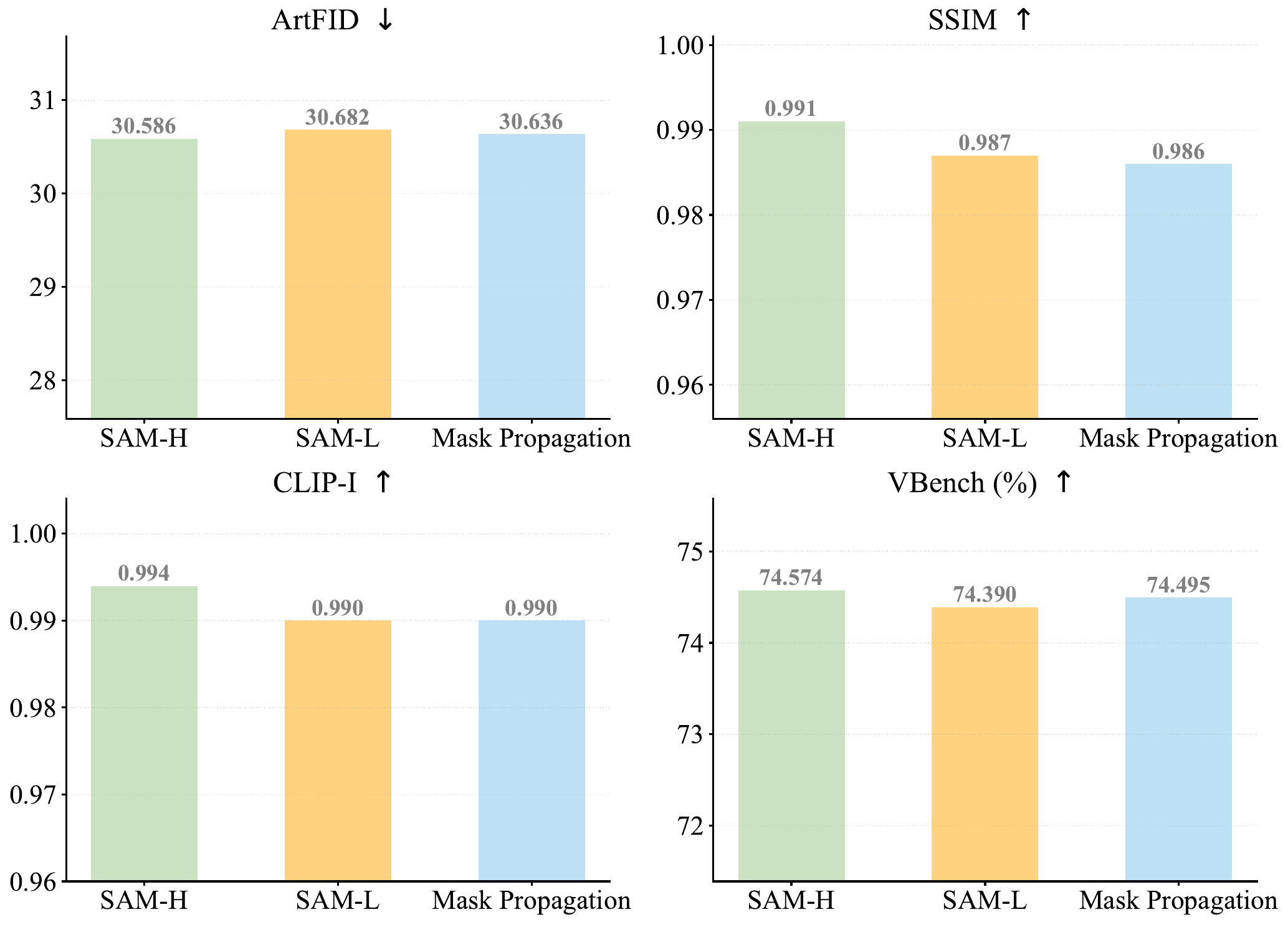}
\caption{
Comparison of stylization performance among our mask propagation strategy, SAM-H, and SAM-L on DAVTG-Laion dataset.
}
\label{fig:ablation_mask_model}
\end{figure}

As shown in Fig.\,\ref{fig:maskpropagation_intro}, our mask propagation strategy demonstrates spatial and temporal advantages over frame-by-frame segmentation with the SAM model~\cite{SAM}.
To further assess the stylization performance of our mask propagation strategy, we perform local video style transfer using masks generated by three approaches: SAM-H, SAM-L, and our mask propagation. This ablation study is performed on the DAVTG-Laion dataset, and the corresponding quantitative comparison is presented in Fig.\,\ref{fig:ablation_mask_model}.
Our mask propagation strategy outperforms SAM-L on the VBench metric, while achieving performance comparable to both SAM-L and SAM-H on the other three metrics, ArtFID, SSIM, and CLIP-I. These results highlight the robustness of our strategy in balancing content and style, preserving local stylization, and maintaining overall visual quality during the localized video style transfer process.

\begin{figure}[!t]
\includegraphics[width=0.95\columnwidth]{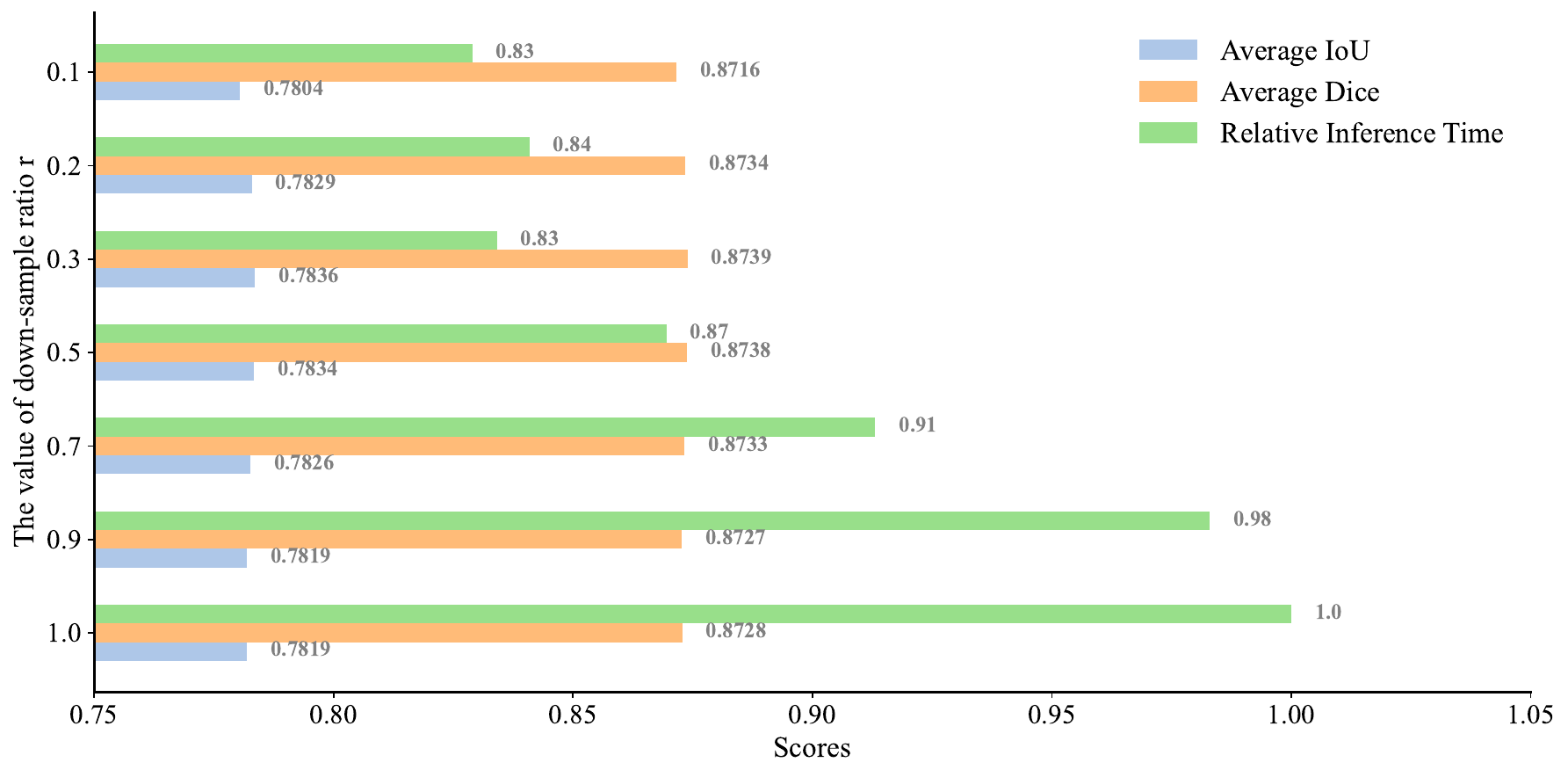}
\caption{Quantitative results of mask propagation with different sampling ratios $r$ on DAVIS~\cite{davis} dataset. Best viewed with zooming-in.}
\label{fig:comparision of downsample}
\end{figure}

In the following section, we conduct an in-depth study of the three proposed modules on the DAVIS-Laion dataset to further substantiate their individual contributions, respectively.

\subsubsection{Analysis of Point-Matching Mask Propagation}
\label{sec:ablation_mask_propagation}
We first investigate the effect of down-sampling rates on propagation accuracy and inference time using the DAVIS dataset~\cite{davis}. Detailed results are presented in Fig.\,\ref{fig:comparision of downsample}. As the down-sampling rate $r$ decreases, the efficiency of our mask propagation improves significantly, while maintaining the accuracy of the final propagation results.
Next, we also examine the impact of anchor frame selection on propagation accuracy and efficiency. As shown in Table\,\ref{tab:comparison_with_different_anchor_frames}, increasing the number of anchor frames improves propagation accuracy but also leads to longer propagation times. Moreover, we observe that as the number of frames increases, the accuracy gains diminish. After considering these trade-offs, we choose the first frame and the previous $n=9$ frames as the final hyper-parameter configuration.
We also compare the results for different values of the hyper-parameter $k$ in $k$-NN~\cite{KNN}. As shown in Table\,\ref{tab:comparison_with_different_k}, the highest propagation accuracy is consistently achieved when $k=15$, without sacrificing the overall propagation efficiency.

\begin{table}[!t]
    \centering
    \caption{Accuracy and efficiency of mask propagation under different values of $n$ on DAVIS-Laion dataset, \textbf{Bold} fonts represent final configuration.
    }
    \begin{tabular}{l c c c}
        \toprule
        Configuration         & Avg IoU$\uparrow$  & Avg Dice$\uparrow$ & Inference Time (s)$\downarrow$ \\ 
        \midrule
        n=1                 & 0.5321  & 0.6505   & 2.365        \\ 
        n=1 + First           & 0.7799  & 0.8714   & 2.423 ($\downarrow$2.5\%)         \\ 
        n=3 + First           & 0.7792  & 0.8711   & 2.477 ($\downarrow$4.7\%)         \\ 
        n=5 + First           & 0.7800  & 0.8716   & 2.577 ($\downarrow$9.0\%)         \\ 
        n=7 + First           & 0.7809  & 0.8721   & 2.673 ($\downarrow$13.0\%)         \\ 
        \textbf{n=9 + First}  & \textbf{0.7816}  & \textbf{0.8726} & \textbf{2.742 ($\downarrow$15.9\%)}    \\ 
        n=11 + First          & 0.7820  & 0.8728   & 2.868 ($\downarrow$21.3\%)         \\ 
        n=13 + First          & 0.7820  & 0.8728   & 2.873 ($\downarrow$21.5\%)         \\ 
        n=15 + First          & 0.7820  & 0.8729   & 2.878 ($\downarrow$21.7\%)       \\ 
        \bottomrule
    \end{tabular}
    \label{tab:comparison_with_different_anchor_frames}
\end{table}

\begin{table}[!t]
    \centering
    \caption{Accuracy and efficiency of mask propagation with
different values of $k$ on DAVIS-Laion dataset. \textbf{Bold} fonts represent final configuration.}
    \begin{tabular}{l c c c}
        \toprule
        $k$ & Avg IoU$\uparrow$ & Avg Dice$\uparrow$ & Inference Time (s)$\downarrow$ \\ 
        \midrule
        40            & 0.7806          & 0.8719          & 2.467          \\ 
        30            & 0.7836          & 0.8739          & 2.458          \\ 
        20            & 0.7836          & 0.8750          & 2.455          \\ 
        \textbf{15}   & \textbf{0.7885} & \textbf{0.8772} & \textbf{2.452} \\ 
        10            & 0.7875          & 0.8766          & 2.452          \\ 
        \bottomrule
    \end{tabular}
    \label{tab:comparison_with_different_k}
\end{table}

\begin{table}[!t]
    \centering
    \caption{
    Comparison of our method with other training-free stylization methods on DAVIS-Laion dataset.
    \textbf{Bold} values indicate final configuration.
    }
    \begin{tabular}{l c c c c} 
        \toprule
        Methods & K-V Repl. & K-V AdaIN & Ours ($\beta_t \uparrow$) & \textbf{Ours ($\beta_t \downarrow$)} \\
        \midrule
        ArtFID$\downarrow$  & 32.561 & 32.581 & 32.158 & \textbf{30.658} \\
        FID$\downarrow$  & 21.389 & 21.829 & 21.620 & \textbf{20.529} \\
        LPIPS$\downarrow$  & 0.454 & 0.427 & 0.422 & \textbf{0.424} \\
        \bottomrule
    \end{tabular}
    \label{tab:attention_shift_comparison}
\end{table}

\subsubsection{Analysis of AdaIN-Guided Localized Video Stylization}
\label{sec:ablation_localized_video_stylization}
We first investigate the effectiveness of our proposed AdaIN-guided attention-shift mechanism and the impact of the monotonicity of the time coefficient $\beta_t$. Specifically, we compare the stylized performance of four configurations: K-V replacement~\cite{StyleID}, K-V AdaIN, our method with increasing time coefficients, and our method with decreasing time coefficients. As shown in Table\,\ref{tab:attention_shift_comparison}, using a decreasing time coefficient enables the AdaIN-guided attention-shift to outperform other training-free methods, such as K-V replacement, K-V AdaIN, or attention-shift with increasing time coefficients.

\begin{table*}[!t]
    \centering
    \caption{Quantitative results of AdaIN-guided attention-shift with different timestep intervals on DAVIS-Laion dataset. 
    \textbf{Bold} fonts indicate final configuration.
    }
    \begin{tabular}{l ccccc ccccc} 
        \toprule
         Timestep Intervals & $\leq 0.6T$ & $\leq 0.5T$ & $\leq 0.4T$ & $\leq 0.3T$ & $\leq 0.2T$ & $\geq \boldsymbol{0.4T}$ & $\geq 0.5T$ & $\geq 0.6T$ & $\geq 0.7T$ & $\geq 0.8T$ \\ 
        \midrule
        ArtFID$\downarrow$ & 31.843 & 32.520 & 33.298 & 34.912 & 35.655 & \textbf{30.658} & 30.835 & 30.868 & 31.019 & 31.893 \\ 
        FID$\downarrow$    & 21.806 & 22.118 & 22.392 & 23.163 & 23.356 & \textbf{20.529} & 20.704 & 20.761 & 20.845 & 21.324 \\ 
        LPIPS$\downarrow$  & 0.396  & 0.407  & 0.424  & 0.445  & 0.464  & \textbf{0.424}  & 0.421  & 0.419  & 0.420  & 0.429  \\ 
        \bottomrule
    \end{tabular}
    \label{tab:attention_shift}
\end{table*}

\begin{table*}[!t]
    \centering
    \caption{
    Quantitative results of AdaIN-guided latent-shift with different timestep intervals on DAVIS-Laion dataset. 
    \textbf{Bold} fonts indicate final configuration.
    }
    \begin{tabular}{l cccccc} 
        \toprule
        Timestep Intervals & $\boldsymbol{0.1T} \sim \boldsymbol{0.2T}$ & $0.2T \sim 0.3T$ & $0.4T \sim 0.5T$ & $0.5T \sim 0.6T$ & $0.7T \sim 0.8T$ & $0.8T \sim 0.9T$ \\  
        \midrule
        ArtFID$\downarrow$   & \textbf{31.925}  & 32.288   & 32.637   & 32.830  & 33.097   & 32.955   \\
        FID$\downarrow$      & \textbf{21.722}  & 21.745   & 21.854   & 22.075  & 22.157   & 22.206   \\
        LPIPS$\downarrow$    & \textbf{0.405}   & 0.419    & 0.428    & 0.423   & 0.429    & 0.420    \\  
        \bottomrule
    \end{tabular}
    \label{tab:latent_shift}
\end{table*}

\begin{figure*}[!t]
\centering
\includegraphics[width=0.95\linewidth]{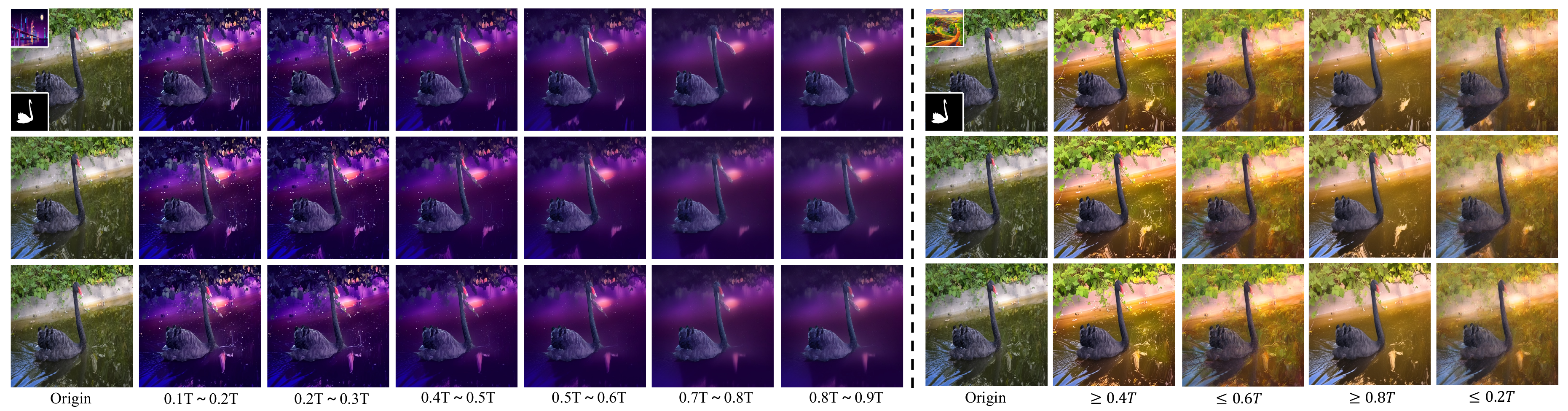}
\caption{
Visual results of latent-shift and attention-shift with different timestep intervals on DAVIS-Laion dataset. Best viewed with zooming-in.
}
\label{fig:latent-attention-shift_with_t}
\end{figure*}

In addition, we further investigate the optimal timestep intervals for AdaIN-guided attention-shift and latent-shift.
As shown in Table\,\ref{tab:attention_shift}, applying attention-shift in the early and middle stages of denoising yields better results.
In contrast, Table\,\ref{tab:latent_shift} shows that latent-shift performs better when applied in the late stages.
Moreover, we present visual results of these ablation experiments, as shown in Fig.\,\ref{fig:latent-attention-shift_with_t}. For AdaIN-guided latent-shift, applying it in the early stages of denoising can cause issues such as content blurriness and semantic loss during the transfer process. In contrast, AdaIN-guided attention-shift produces better visual results when applied in the early stages of denoising than in the later stages.

\begin{figure}[t]
\centering
\includegraphics[width=0.95\columnwidth]{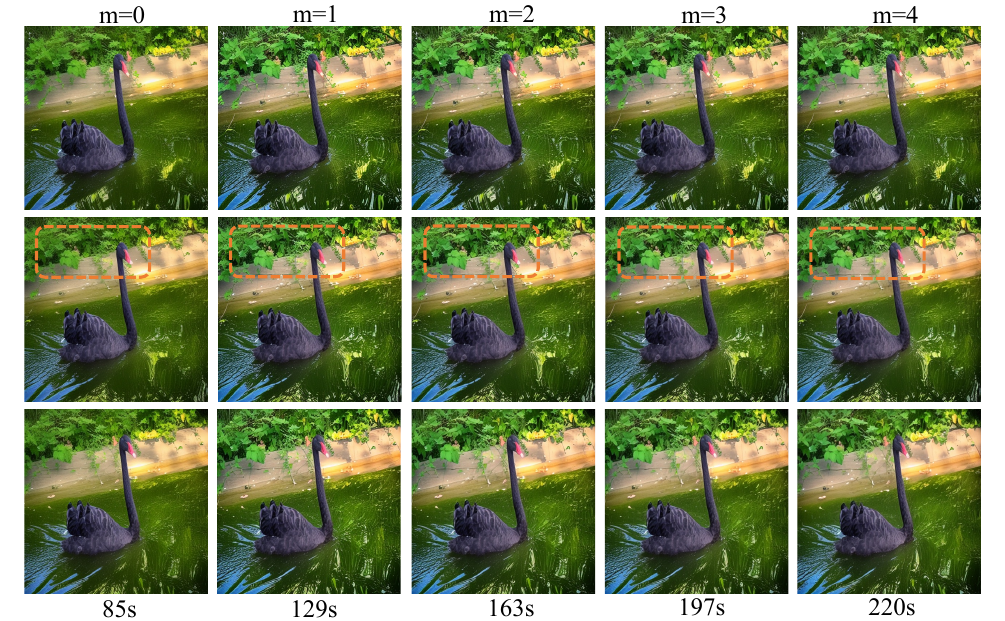}
\caption{Visual results of sliding-window consistent smoothing with different window sizes on DAVIS-Laion dataset. Best viewed with zooming-in.}
\label{fig:ablation of slide windows with window size}
\end{figure}

\subsubsection{Analysis of Sliding-Window Consistency Smoothing}
\begin{figure}[!t]
\centering
\includegraphics[width=0.95\columnwidth]{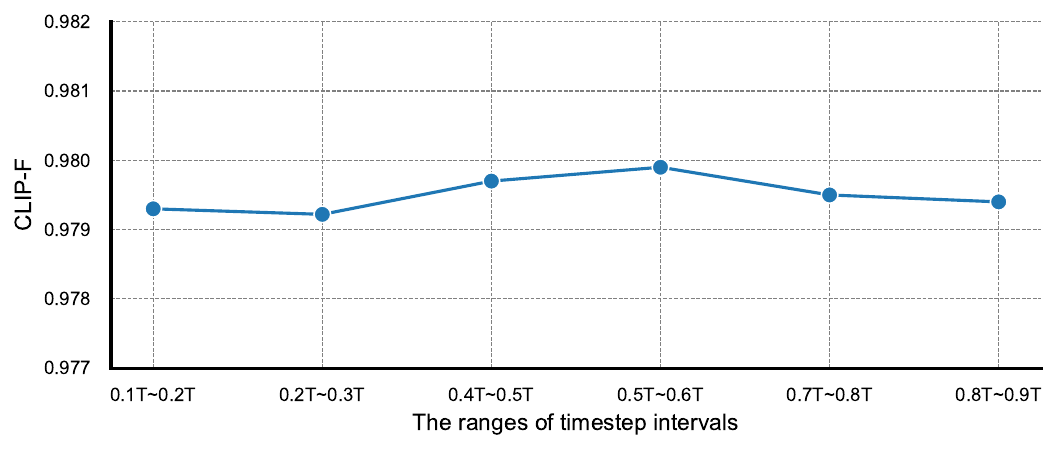}
\vspace{-0.5em}
\caption{Quantitative results of sliding-window consistent smoothing with varying timestep intervals. Best viewed with zooming-in.}
\label{fig:ablation of slide windows with timesteps}
\end{figure}

To assess the impact of different window sizes $m$ on stylistic effects and computational costs in the sliding-window consistency smoothing scheme, we present the visual effects and corresponding inference time for various window sizes in Fig.\,\ref{fig:ablation of slide windows with window size}. While larger window sizes yield slight visual improvements, they also result in significant inference costs. Ultimately, we select $m=2$ to balance stylistic effects and computational costs.
Furthermore, we also explore how applying this strategy at different timestep intervals influences the final temporal consistency. As shown in Fig.\,\ref{fig:ablation of slide windows with timesteps}, we observe that the effectiveness of the smoothing strategy varies across the early, middle, and late stages of the denoising process, with the best results achieved during the middle stage.

\subsection{Addtional Analysis}
\label{sec:additional_analysis}
In this section, we further analyze the \textit{bottleneck of inference time}, the \textit{constraints in mask propagation}, and the \textit{impacts of inversion}.

\begin{figure}[!t]
\centering
\includegraphics[width=0.95\columnwidth]{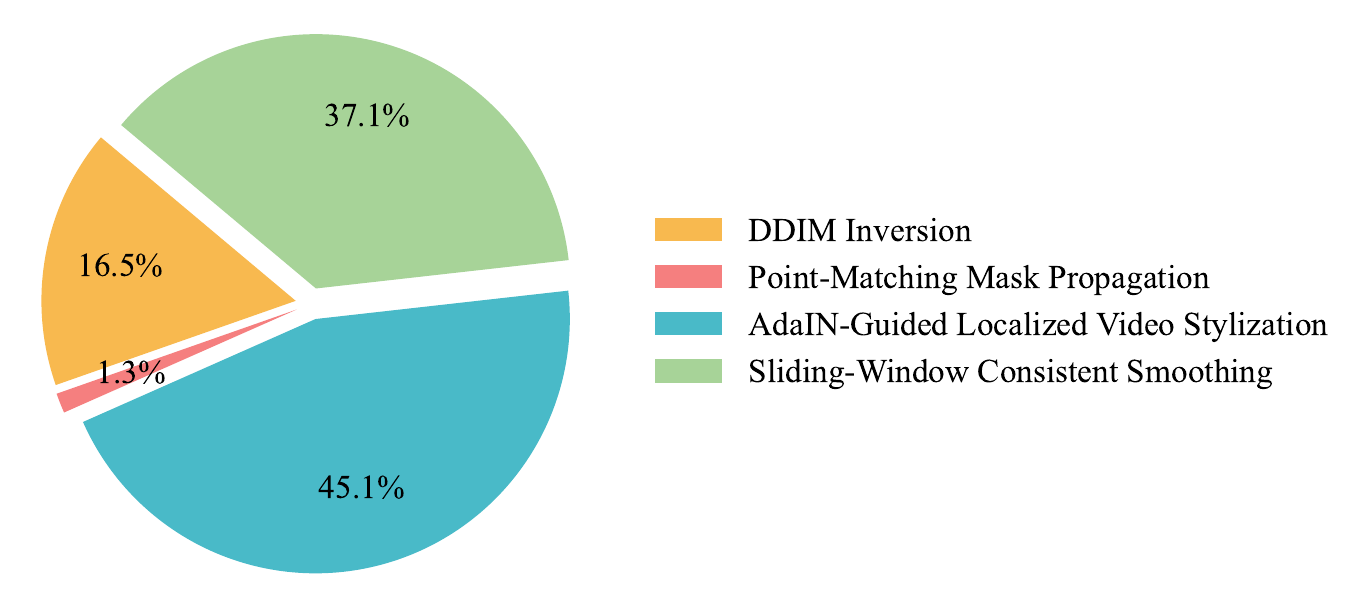}
\caption{
Inference time distribution of different components in UniVST.
Sliding-window consistent smoothing exhibits an inference bottleneck.
}
\vspace{-1.5em}
\label{fig:time_analysis}
\end{figure}

\textbf{Bottleneck of Inference Time.}
In Table\,\ref{tab:comparison_all}, our proposed UniVST exhibits the longest inference time among all compared methods, revealing a key computational bottleneck.
To better understand this limitation, we perform a detailed breakdown of the inference time distribution across the core modules of UniVST, as illustrated in Fig.\,\ref{fig:time_analysis}. These modules include: \textit{DDIM Inversion}, \textit{Point-Matching Mask Propagation}, \textit{AdaIN-Guided Localized Video Stylization}, and \textit{Sliding-Window Consistent Smoothing}.
Our analysis indicates that the sliding-window consistent smoothing module accounts for about 37.1\% of the total overhead, primarily due to the dynamic estimation of inter-frame optical flow within each sliding window during specific denoising intervals. 
This component involves a trade-off: although it enhances temporal consistency in a training-free manner during the denoising process, the frequent estimation of optical flow inevitably introduces additional computational overhead.
To address the above trade-off, we propose two training-free alternatives: (1) employing a 3D VAE decoder to reconstruct temporally coherent video from the denoised latent representation; and (2) incorporating a plug-and-play pretrained adapter after the self-attention module for temporal modeling, similar to the motion module in AnimateDiff~\cite{AnimateDiff}.

\begin{figure}[!t]
\centering
\includegraphics[width=0.95\linewidth]{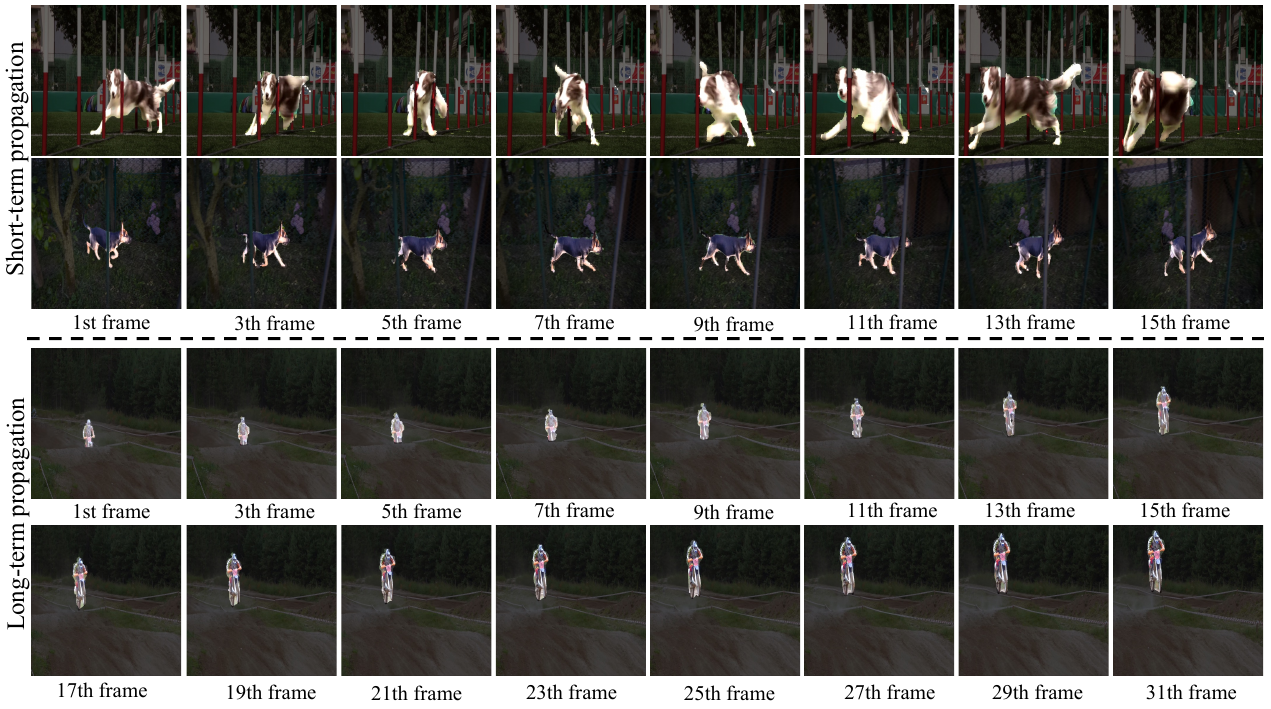}
\caption{
Visual results of our point-matching mask propagation strategy on the DAVIS dataset~\cite{davis}. The masks are blended with the original video to emphasize the foreground, while dimming the background. Our propagation strategy can effectively adapt to various complex scenarios.
}
\label{fig:result_of_mask}
\end{figure}

\textbf{Constraints in Mask Propagation.}
We present qualitative examples of our mask propagation results in Fig.\,\ref{fig:result_of_mask}.
In the short-term propagation setting, we evaluate scenarios involving fast-moving foreground objects, deformations, and occlusions. Despite these complexities, our method demonstrates precise object tracking and robust mask alignment, consistently preserving accuracy across diverse conditions.
For long-term propagation, our approach continues to deliver strong performance. The integration of anchor frames enables stable and accurate mask propagation over extended temporal distances, mitigating typical degradation issues associated with long sequences.
Although our point-matching mask propagation strategy enables zero-shot object localization, it presents several limitations:
\textit{1. Challenges in Multi-Object Scenarios.} In scenes containing multiple similar objects (\emph{e.g.}, two individuals), the semantic similarity between instances can lead to confusion during propagation. This often results in incorrect mask associations, particularly as the propagation distance increases, causing error accumulation over time. While our current evaluation focuses on diverse yet single-object scenes—where such issues have minimal impact—extending the method to robustly handle multi-object scenarios remains an important direction for future work.
\textit{2. Limitations in Fine-Grained Mask Propagation.} Although the propagation achieves an average Dice score of approximately $0.87$, it occasionally fails to capture fine object boundaries, leading to incomplete or imprecise tracking.
As shown in Fig.\,\ref{fig:result_of_mask}, while the method effectively handles fast motion and occlusions, it struggles to retain detailed edge information. Nonetheless, our localized latent blending strategy, which performs dynamic region-wise fusion during the denoising phase, effectively mitigates this limitation by reducing reliance on high-precision masks.

\begin{figure}[!t]
\centering
\includegraphics[width=0.95\linewidth]{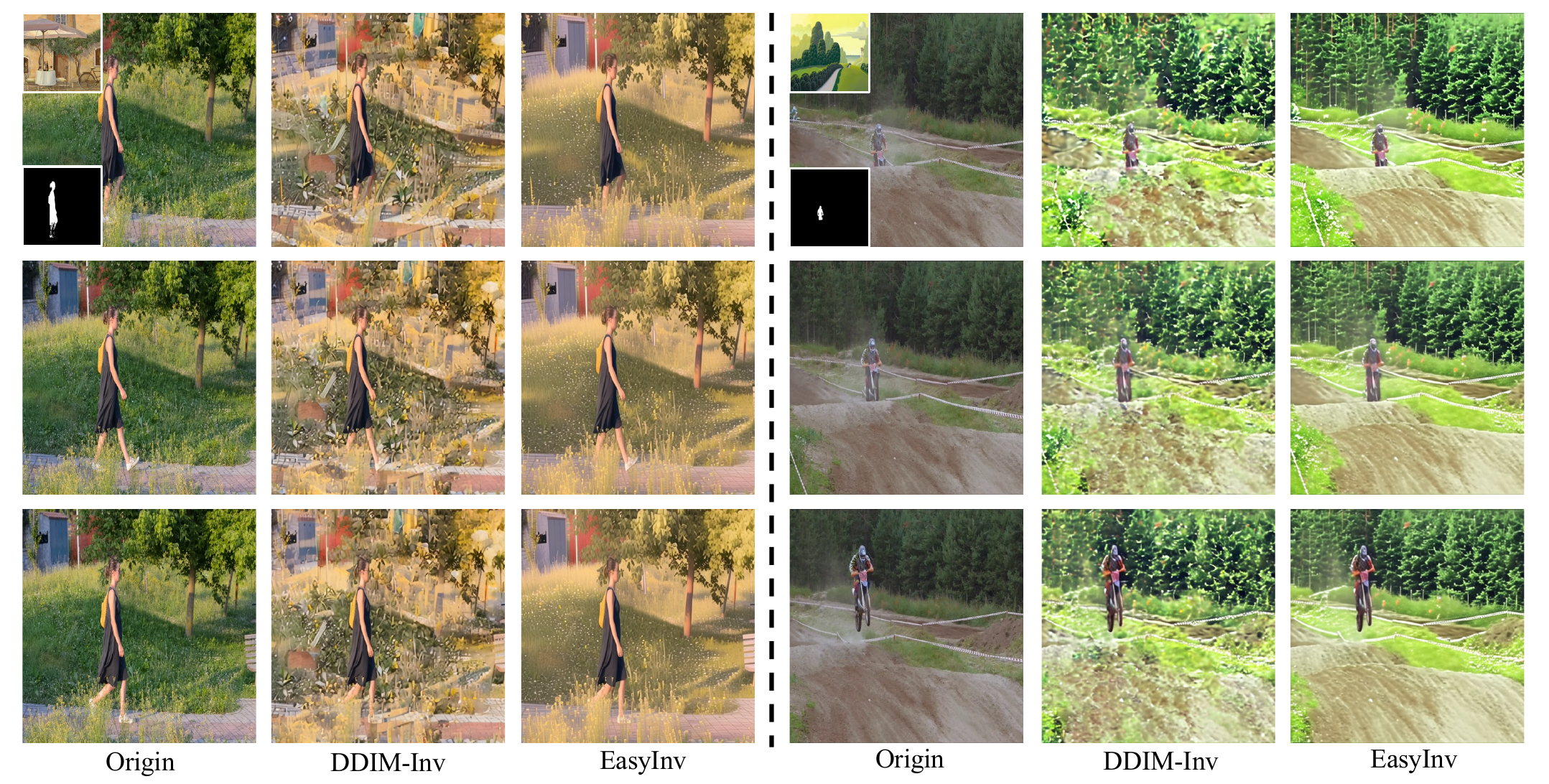}
\caption{Qualitative comparisons of different inversion methods. EasyInv~\cite{easy_inv} demonstrates better transfer quality compared to DDIM-Inv~\cite{DDIM}.
}
\label{fig:inversion_comparison}
\end{figure}

\textbf{Impacts of Inversion.}
As a training-free stylization method, UniVST relies on the initial noise derived from the original video, thereby rendering the inversion step crucial. The overall performance of UniVST is somewhat sensitive to the quality of the inverted noise, which serves as a critical foundation for subsequent stylization. Inadequate inversion can compromise both content fidelity and stylistic consistency.
As illustrated in Fig.\,\ref{fig:inversion_comparison}, we observe that naive inversion methods often fail to preserve fine-grained semantic details and adversely impact subsequent mask propagation. To mitigate the artifacts introduced by imperfect inversion, we employ EasyInv~\cite{easy_inv}, a more advanced inversion technique. By leveraging EasyInv, we achieve higher-fidelity noise reconstructions, which in turn lead to improved visual quality and stable mask propagation throughout the stylization process.

\subsection{Addtional Comparison}
\begin{figure*}[!t]
\centering
\includegraphics[width=0.95\linewidth]{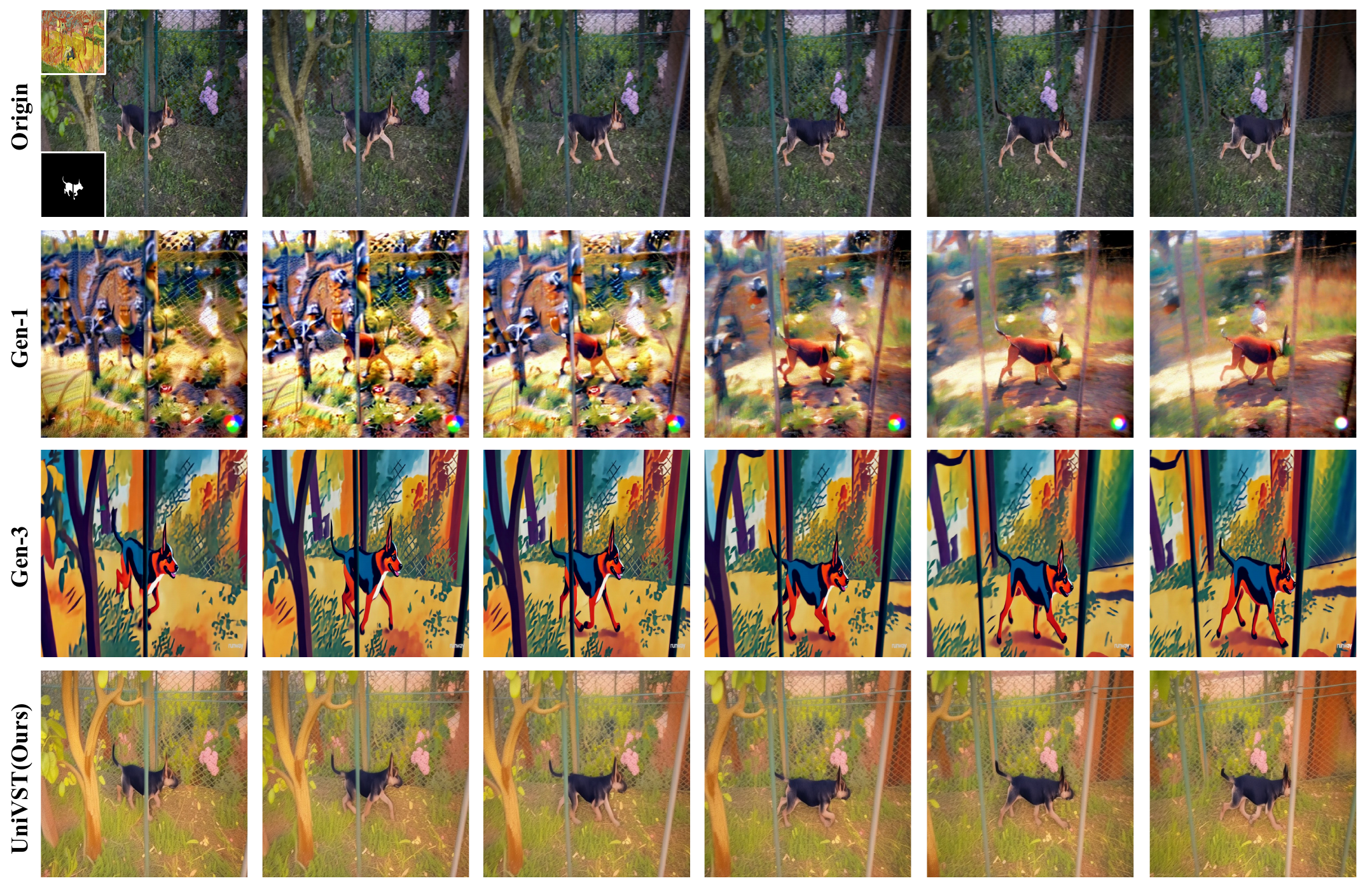}
\caption{Qualitative comparisons of our proposed UniVST with existing commercial models. 
The first row presents the original video, reference image, and mask, followed by the results of Gen-1~\cite{Gen-1} in the second row, Gen-3 in the third row, and our proposed UniVST in the final row.
}
\vspace{-1.0em}
\label{fig:comparison_with_commercial}
\end{figure*}

Furthermore, we have also compared our UniVST with existing commercial models, namely Gen-1~\cite{Gen-1} and Gen-3, as illustrated in Fig.\,\ref{fig:comparison_with_commercial}. 
UniVST demonstrates clear superiority in two critical aspects: achieving a closer style similarity to the reference image and maintaining higher content fidelity to the original video. In addition, UniVST uniquely excels in accurately preserving the style consistency of local objects, thereby ensuring a more cohesive and visually appealing result, even in complex or dynamic scenes.

\subsection{Multi-Step Video Style Transfer}
\begin{figure}[!t]
\centering
\includegraphics[width=0.95\linewidth]{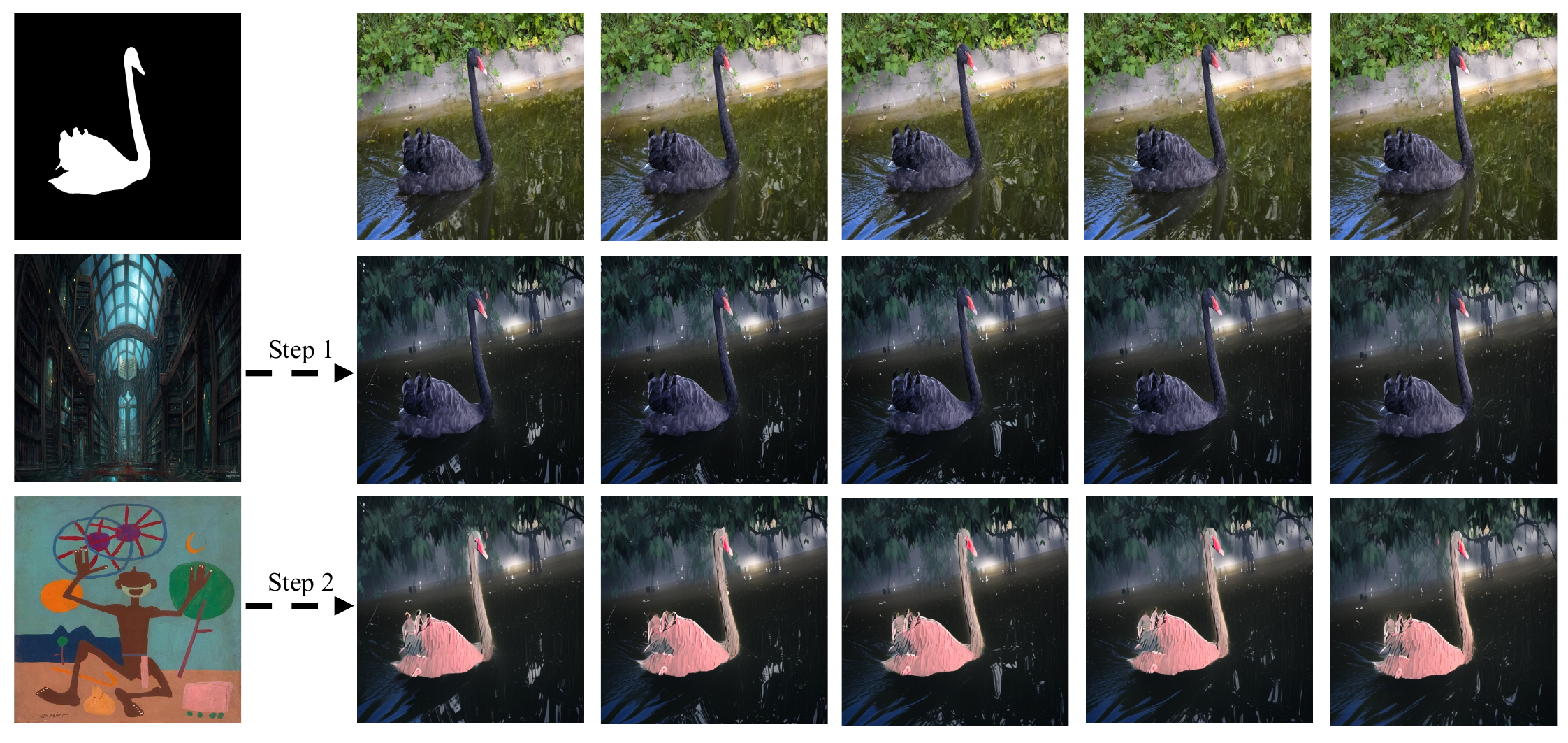}
\caption{
Multi-step video style transfer: background is first stylized, and then the foreground is stylized based on the transformed videos above.
}
\vspace{-1.0em}
\label{fig:multi_transforms1}
\end{figure}

Apart from the cases presented in the experimental section, UniVST showcases its versatility by enabling multi-step video style transfer, which allows for more customized transformations and the creative potential of video editing.
For instance, users can start by stylizing the background of the video using a mask to isolate it from the rest of the scene. After completing this step, they can invert the mask to focus on the previously excluded foreground and apply a separate style transfer to it. This sequential process not only provides greater control but also enables creative layering of styles. 
As illustrated in Fig.\,\ref{fig:multi_transforms1}, employing distinct style images for the background and foreground in these two stages produces a visually striking and dynamic effect, enhancing the artistic flexibility of the method.

\section{Limitation and Future Work}
While our UniVST framework shows a clear advantage in localized video style transformation, it also has limitations:
(1) The accuracy of mask propagation and video content fidelity may be significantly affected by inaccuracies in the inversion process.
(2) The style transfer mechanism, guided solely by prior knowledge of LDM~\cite{LDM}, may struggle with certain style images, especially those with nuanced color distributions, leading to suboptimal performance.
(3) The sliding-window smoothing technique, based on optical flow, can introduce additional inference time.
Future research could explore improved stylization techniques to enhance visual effects and reduce inference overhead from optical flow prediction.

\section{Conclusion}
This paper presents UniVST, a unified framework for localized video style transfer without additional training, marking a significant advancement over traditional methods that require full video style transfer. Our contributions include:
(1) A point-matching mask propagation strategy that eliminates the need for tracking models, enabling streamlined style transfer to specific video objects.
(2) A training-free AdaIN-guided localized video stylization mechanism that operates at both the latent and attention levels, ensuring a balance between content fidelity and style richness.
(3) A sliding-window consistent smoothing scheme that uses optical flow to refine noise and update the latent space, improving temporal consistency and reducing artifacts.
Extensive experiments show that our UniVST outperforms existing methods in both qualitative and quantitative evaluations, effectively preserving the style of the primary object while ensuring temporal consistency. Furthermore, it can be extended to a broader range of customized models.

\section*{Acknowledgments}
This work was supported by the National Science Fund for Distinguished Young Scholars (No. 62025603 and No. 62525605), the National Natural Science Foundation of China (No. U21B2037, No. U22B2051, No. U23A20383, No. 62176222, No. 62176223, No. 62176226, No. 62072386, No. 62072387, No. 62072389, No. 62002305, and No. 62272401), and the NUS Start-up Grant (No. A-0010106-00-00).

\bibliographystyle{IEEEtran}
\bibliography{IEEEabrv,reference}

\end{document}